\documentclass[sigconf]{acmart}

\usepackage{booktabs} 
\usepackage{array}
\usepackage{xcolor}
\usepackage{multirow}
\newcolumntype{H}{>{\setbox0=\hbox\bgroup}c<{\egroup}@{}}
\newcommand{\cfbox}[2]{%
	\colorlet{currentcolor}{.}%
	{\color{#1}%
		\setlength{\fboxsep}{1.5pt}\setlength{\fboxrule}{1pt}\fbox{\color{currentcolor}#2}}%
}

\iffalse
\setcopyright{acmcopyright}

\acmDOI{10.1145/3240508.3240624}

\acmISBN{978-1-4503-5665-7/18/10}

\acmConference[MM '18]{2018 ACM Multimedia Conference}{October 22--26, 2018}{Seoul, Republic of Korea}
\acmBooktitle{2018 ACM Multimedia Conference (MM '18), October 22--26, 2018, Seoul, Republic of Korea}
\acmYear{2018}
\copyrightyear{2018}

\acmPrice{15.00}

\acmSubmissionID{559}
\else
\setcopyright{none}
\renewcommand\footnotetextcopyrightpermission[1]{} 
\fi

\usepackage{subcaption,amssymb,amsmath,rotating,fancyhdr,tabularx, capt-of,soul}
\usepackage[algoruled,boxed,lined]{algorithm2e}

\soulregister\cite7
\soulregister\ref7
\soulregister\pageref7

\definecolor{lightblue}{rgb}{.7,1,1}
\definecolor{lightgreen}{rgb}{.8,1,.8}
\definecolor{lightorange}{rgb}{1,0.8,0.3}
\definecolor{darkgreen}{rgb}{0.2,.5,0.2}
\definecolor{darkergreen}{rgb}{0.02,.2,0.05}

\newcommand{\cout}[1]{} 

\newcommand{\todo}[1]{}

\newcommand{\note}[1]{} 

\newcommand{\ie}{\emph{i.e.\;}}
\newcommand{\eg}{\emph{e.g.\;}}
\newcommand{\etal}{\emph{et al.\;}}
\newcommand{\etc}{\emph{etc.\;}}
\newcommand{\cf}{\emph{c.f.\;}}

\soulregister\etal7
\soulregister\ie7
\soulregister\eg7
\soulregister\cf7
\begin{document}
\title{Predicting Visual Context for Unsupervised Event Segmentation \\in Continuous Photo-streams}

\author{Ana Garc\'{i}a del Molino}
\iffalse
\additionalaffiliation{%
	\institution{School of Computer Science and Engineering, Nanyang Technological University, Singapore}
}
\affiliation{\institution{Institute for Infocomm Research, A*STAR, Singapore}}
\email{stugdma@i2r.a-star.edu.sg}
\else
\affiliation{%
	\institution{School of Computer Science and Engineering, Nanyang Technological University, Singapore}
}
\additionalaffiliation{\institution{Institute for Infocomm Research, A*STAR, Singapore}}
\email{ana002@e.ntu.edu.sg}
\fi
\orcid{0000-0002-0779-1656}
\author{Joo-Hwee Lim}
\affiliation{\institution{Institute for Infocomm Research, A*STAR, Singapore}
}
\email{joohwee@i2r.a-star.edu.sg}
\author{Ah-Hwee Tan}
\affiliation{\institution{School of Computer Science and Engineering, Nanyang Technological University, Singapore}
}
\email{asahtan@ntu.edu.sg}
\thanks{This work is supported by A*STAR JCO Grant 1335h00098 (REVIVE), IAF-ICP Grant ICP1600003 (VInspection), and A*STAR Singapore International Graduate Award (SINGA).}
\renewcommand{\shortauthors}{Garc\'{i}a del Molino et al.}

\begin{abstract}
Segmenting video content into events provides semantic structures for indexing, retrieval, and summarization.
Since motion cues are not available in continuous photo-streams, and annotations in lifelogging are scarce and costly, the frames are usually clustered into events by comparing the visual features between them in an unsupervised way.
However, such methodologies are ineffective to deal with heterogeneous events,~\eg taking a walk, and temporary changes in the sight direction,~\eg at a meeting.
To address these limitations, we propose Contextual Event Segmentation (CES), a novel segmentation paradigm that 
uses an LSTM-based generative network to model the photo-stream sequences, predict their visual context, and track their evolution.
CES decides whether a frame is an event boundary by comparing the visual context generated from the frames in the past, to the visual context predicted from the future.
We implemented CES on a new and massive lifelogging dataset consisting of more than $1.5$ million images spanning over $1,723$ days. Experiments on the popular EDUB-Seg dataset show that our model outperforms the state-of-the-art by over $16\%$ in f-measure. 
Furthermore, CES' performance is only $3$ points below that of human annotators.

\end{abstract}


\maketitle

\section{Introduction}
\begin{figure}[t]
	\centering
	\includegraphics[width=.98\linewidth]{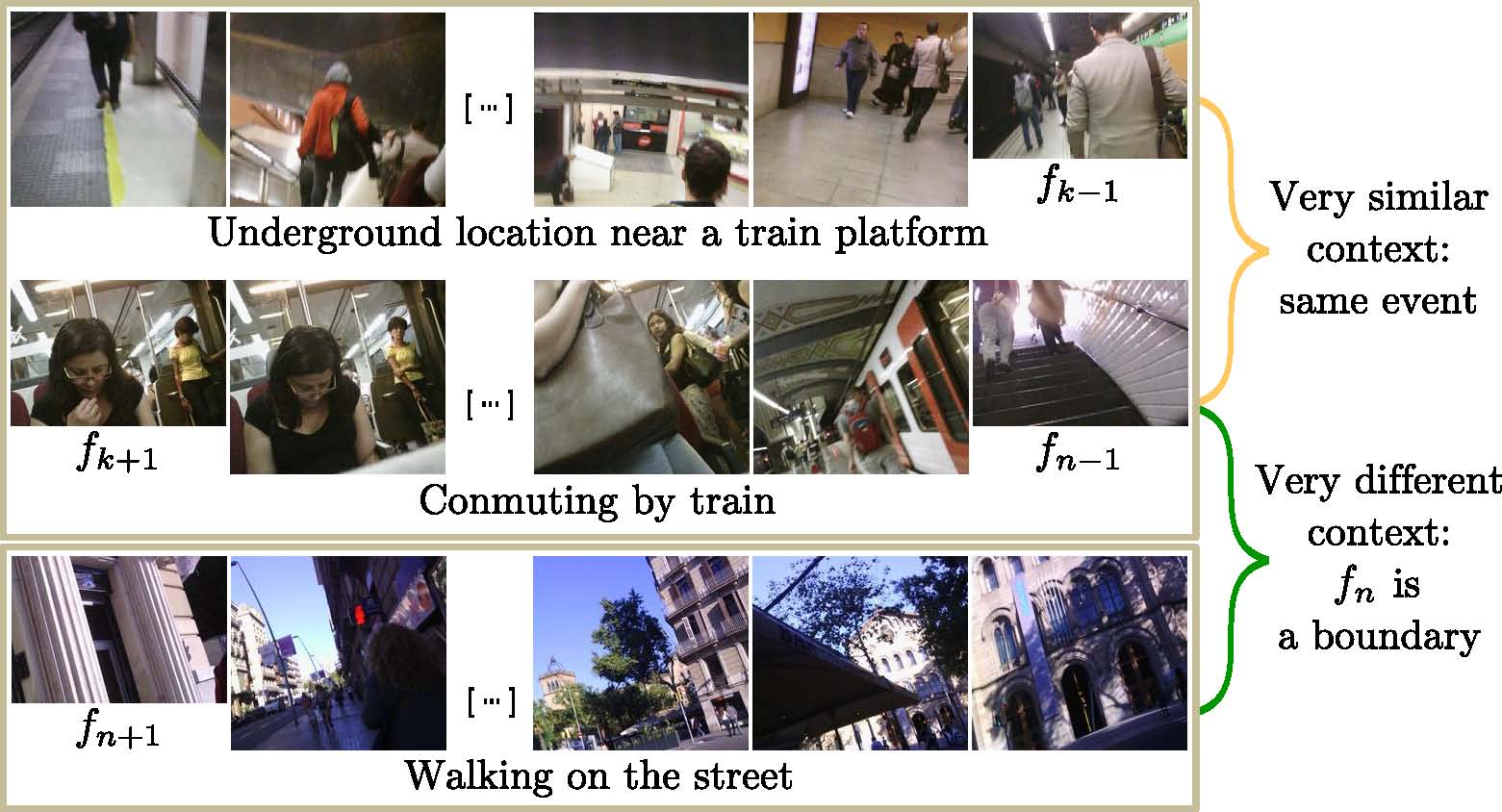}\,%
	\vspace*{-1ex}
	\caption{As humans, we define a new event when the new sequence of frames differs from our understanding of the previous frame sequence. CES models such intuitive framework of perceptual reasoning by predicting the visual context of the photo-stream. At each timestep, it compares the context predicted from the past sequence to the context predicted from the future sequence.}
	\vspace*{-2ex}
	\label{fig:teaser}
\end{figure}
Continuously recording our lives in the form of images can be of great usefulness for memory enhancement, tracking of the activities of daily living, and other related healthcare applications. However, lifelogging has an overload problem both in time and space: Lifelogging cameras take a minimum of $2$ pictures per minute, which can add to more than $1,000$ pictures a day, \ie $100 Gb$ per year. Such vast load of data requires hours of manual analysis to, for example, select your day's highlights, check what you ate and drank the past month, or monitor your grandparent's routines. Hence, automatic tools to extract highlights and life patterns are needed~\cite{EEG2014, harvey2016remembering, XiongSnap, GarciadelMolino2018phd}. However, analyzing lifelogs entails two great challenges related to its Low Time Resolution (LTR) and wearable nature: First, dramatic visual changes between consecutive frames even if these correspond to the same event. Second, a substantial presence of visual occlusions, walls and ceilings in the field of view, and frequent changes of visual orientations.

Extensive research has been conducted to retrieve specific events or obtain summaries from First Person View (FPV) videos~\cite{MySurvey, BolanosSurv}. While event segmentation is needed for a complete, informative and diverse summary that includes most life events in the recording, little work has been done to that effect~\cite{BolanosSurv, NTCIR_db}. Many approaches to segment High Time Resolution (HTR) video use motion between frames to infer the wearer's activity, but motion cues are not available in photo-streams. 
 Furthermore, obtaining annotations for such large datasets is very costly. As such, one can only resort to visual features and sensor metadata, and unsupervised techniques such as K-Means~\cite{LifeLogTask17_CLEF} and probabilistic models~\cite{dimiccoli2017sr}. 
 
Due to these limitations, current automated methods usually fail at modeling the frame sequences. As a consequence, they cannot perceive the overall context in heterogeneous events, and usually misinterpret occlusions and occasional diversions within events as different episodes. Our ambition is to build a segmentation model that mimics the human reasoning, as people can easily detect and discard such noise by comparing the new visual input with their understanding of both the previous and following scene (see Fig.~\ref{fig:teaser}).

In this work, we introduce Contextual Event Segmentation (CES), a novel event segmentation technique that, 
given a sequence of frames, predicts its visual context and then compares it to the context corresponding to the ensuing sequence. An LSTM-based generative model, that we call VCP, is used to predict the visual context. It is able to model our daily activities and learn the associations between different scenes, \eg a train commute will include corridors, stairs, a platform, the interior of a wagon, \etc
To train VCP we introduce \textit{R3}, a novel and vast dataset for unsupervised lifelog analysis. It consists of over $1,5$ million images that depict the daily activities of $57$ different users over a total of $1,723$ days.

The main contributions of this paper are:
\begin{itemize}
\item[(i)] a segmentation approach that mirrors the human perceptual reasoning when segmenting photo-streams into events. In a series of experiments, CES proves to be superior to the state of the art by over $16\%$ in f-measure, and even competitive against manual annotations.
\item[(ii)] an LSTM-based generative model to predict visual context from a sequence of frames. We observe that the model learns event traits in common daily activities.
\item[(iii)] a large-scale lifelogging dataset containing $1,500,890$ images from $57$ users. To our knowledge, R3 is the largest FPV dataset currently available\footnote{The data is accessible from http://dx.doi.org/10.17632/ktps5my69g.1}.
\end{itemize}

\section{Related Work}
FPV content entails three main challenges: its unconstrained nature, its continuous stream of consecutive events, and its poor visual quality.
In particular, the purpose of lifelogging is to have a diary of our lives. However, such huge amount of visual content must be summarized to be of practical use.
The summary of these photo-streams should be complete, informative and diverse. When no query is given to constraint the content of the summary, the maximum variety of events should be included. 
To do so, the content must first be segmented into \emph{subshots} in the case of High Temporal Resolution (HTR) videos,  or \emph{events} in the case of Low Temporal Resolution (LTR) videos (or photostreams).

\paragraph{Temporal segmentation in High Temporal Resolution First Person View}
Third Person View (TPV) event segmentation approaches typically identify shot boundaries by detecting abrupt changes between consecutive frames~\cite{MoneySurv,HuSurv}. However, FPV content is not comprised of separate shots, but rather a succession of events with smooth transitions, where event boundaries are not well defined.

Most FPV approaches for event segmentation use motion cues, both visual (\eg optical flow, blurriness)~\cite{AizSum01,NgSum, LuSD,GygliSum14,VariniPref, PolegSeg, PolegCNN, AVS} and from sensors~\cite{AizSum01,spriggs2009temporal}. Such features are used to predict the wearer's activity or attitude patterns using probabilistic models~\cite{VariniPref} and deep learning~\cite{PolegCNN}, to segment the videos accordingly.
Other methods resort to visual similarity between groups of frames (\eg color, GIST, CNN hash)~\cite{LeeDisco12,XiongSnap,LeePred15,AizSum01,NgSum,LuSD,bettadapura2016picturesque,XuGaze,PotapovSum}. Temporally constrained clustering~\cite{LeePred15} and statistical frameworks~\cite{PotapovSum} have been used to determine whether the visual differences correspond to event boundaries or just abrupt head movements.

 

\paragraph{Temporal segmentation in Low Temporal Resolution First Person View}
In the case of lifelog photo-streams, frames can be up to $30$ seconds apart. In such low temporal resolution, content may change a lot between consecutive frames even if they are part of the same event, and as Bolanos~\etal remark in~\cite{BolanosSurv}, visual motion information is unavailable (sensor information may sometimes be available~\cite{DohertyLL08}). 
Given the limited amount of annotated data, event segmentation is very often unsupervised, performed via K-Means and other hierarchical clustering algorithms on visual cues (\eg color, CNN hashes)~\cite{lin2006structuring, DohertyLL08,DohertyRet08,bolanos2015visual, I2R2017CLEF, I2R2017NTCIR, yamamoto2017pbg}. An exception to these unsupervised methods is ~\cite{furnari2018personal}, in which a personal location classifier is trained for each user, and events are segmented according to changes in the wearer's location. Since these methods often ignore the semantic nature of the frames, Dimiccoli~\etal~\cite{dimiccoli2017sr} propose defining the frames with semantic and contextual cues defined by CNN features and linguistic information. The relation between frames is assessed using a WordNet~\cite{WordNet} based knowledge graph, and the event boundaries are found using a graph-cut algorithm integrating an agglomerative clustering. Such segmentation methodology relies on the cross-analysis of consecutive frames, and cannot detect change points between two events with heterogeneous visual content, nor ignore small and isolated visual changes within an event.

To address this limitation, we present a novel event segmentation paradigm in which each frame is understood as part of a global sequence. As such, the visual context of the upcoming frame can be predicted from the preceding sequence of frames. This prevents the model from detecting false positives due to abrupt changes between consecutive frames, and allows it to understand the nature of heterogeneous events.


\paragraph{Sequence embedding for photo-album summarization and activity classification}
Addressing the problem of story-telling from albums of $10$ to $50$ photos, Yu~\etal~\cite{yu2017hierarchically} use a Recurrent
Neural Net (RNN) to encode the local album context for each photo, so that the best key-frames can be selected. Liu~\etal~\cite{liu2017let} 
use Gated Recurrent Units (GRUs) to align the local storylines into the global sequential timeline. To obtain better event descriptions, they further leverage the semantic coherence in a photo stream by jointly embedding the images and sentences into a common semantic space.

Using video content, Bhatnagar~\etal~\cite{bhatnagarunsupervised} obtain good results at describing egocentric motor actions (\eg \textit{stir}, \textit{fold}, \textit{open}) in HTR videos using an hybrid CNN-LSTM auto-encoder. 
Similarly, Srivastava~\etal~\cite{srivastava2015unsupervised} learn spatio-temporal features using a sequence-to-sequence future prediction model, proving that such an architecture is more efficient than an auto-encoder.

Whereas both \cite{bhatnagarunsupervised, srivastava2015unsupervised} learn the spatio-temporal features from the raw frames in HTR video, we propose learning a global semantic visual context from the visual features of LTR frame sequences.


\section{Contextual Event Segmentation}
\subsection{Overview}
Given a continuous stream of photos, we, as humans, would identify the start of an event if the new frame differs from our expectation of what should follow the preceding sequence. We would also check whether that frame is consistent with the subsequent image sequence (or scene). If the new scene spans a very short time and returns to the previous, we would ignore it as an extra event, but rather wrap it within the current event (\eg going for a bottle of water while watching TV). Therefore, we would frequently look forward and backward to verify whether it was a new event, or just a brief diversion or local outlier.


The proposed model is analogous to such intuitive framework of perceptual reasoning. It uses an encoder-decoder architecture to predict the visual context at time $t$ given the images seen before, \ie the past. A second visual context is predicted from the ensuing frames, \ie the future. If the two predicted visual contexts differ greatly, CES will infer that the two sequences (past and future) correspond to different events, and will consider $frame_t$ as a candidate event boundary.

Therefore, CES consists of two modules (\cf Algorithm~\ref{al:Alg}): First, the Visual Context Predictor (VCP), that predicts the visual context of the upcoming frame, either in the past or in the future depending on the sequence ordering. Second, the event boundary detector, that compares the visual context at each time-step given the frame sequence from the past, with the visual context given the sequence in the future.

\subsection{Visual Context Predictor}

Inspired by \cite{bhatnagarunsupervised, srivastava2015unsupervised, yu2017hierarchically}, we propose predicting the visual context from a sequence of frames with a Long-Short Term Memory network. LSTM networks are a type of Recurrent Neural Network that learn long-time dependencies through four hidden layers, \ie the gates. Thus, LSTMs can aggregate the information they receive by learning to forget. Their mathematical formulation can be expressed as

\begin{equation}
\label{eq:LSTM}
\begin{split}
\left( \begin{tabular}{@{}c@{}}    \underline{i} \\
          \underline{f} \\
          \underline{o} \\
          \underline{g} \end{tabular}
         \right)&  = \left(\begin{tabular}{@{}c@{}} \textit{sigm} \\
          \textit{sigm}\\
          \textit{sigm} \\
          \textit{tanh} \end{tabular}
         \right) \left(\begin{bmatrix}
           \mathbf{x_t} \\
           \mathbf{h_{t-1}}
         \end{bmatrix}\mathbf{W}
         \right) \\
\mathbf{c_t} & = \underline{f} \circ \mathbf{c_{t-1}} + \underline{i} \circ \underline{g} \\
\mathbf{h_t} & = \underline{o} \circ tanh(\mathbf{c_t}) \, ,
\end{split}
\end{equation}

\noindent where $\underline{i}$, $\underline{f}$, $\underline{o}$, and $\underline{g}$ correspond to the four gates of the unit (\textit{input}, \textit{forget}, \textit{output}, and \textit{input modulation}), $\circ$ is
the element-wise product, $\mathbf{W}$ are the network weights~\cite{gal2016theoretically}, and $\mathbf{c_t}$ and $\mathbf{h_t}$ are the cell state and the hidden state, respectively, at time-step $t$.

\begin{figure}[t!]
	\centering
	\includegraphics[width=.99\linewidth]{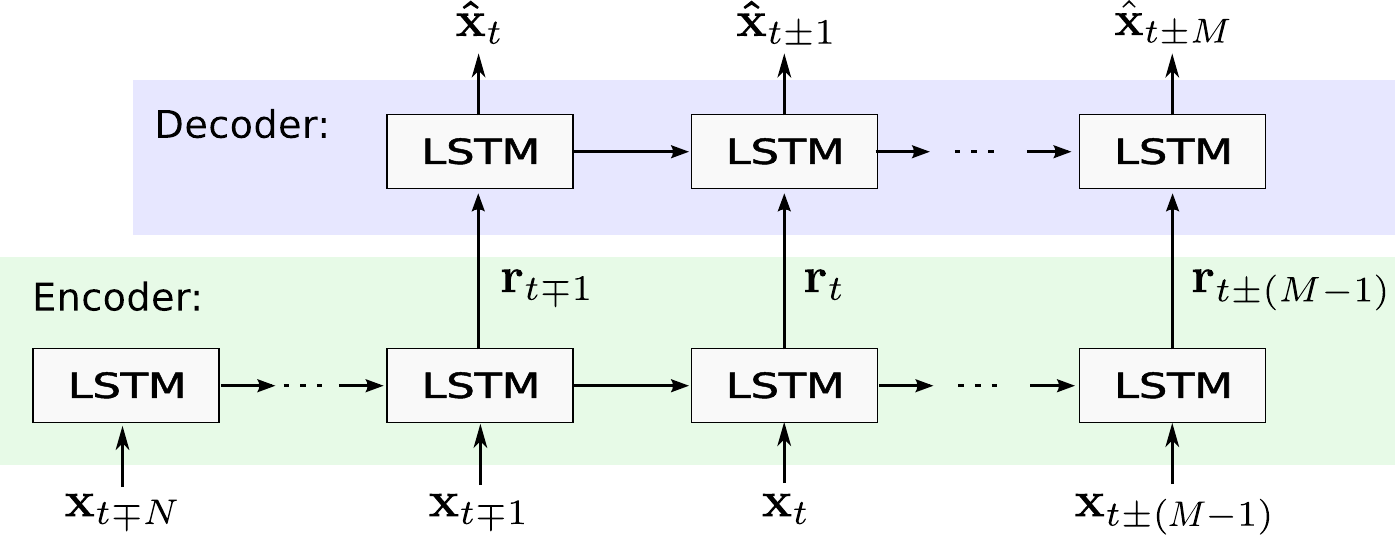}
	\vspace*{-5pt}
	\caption{Training of the Visual Context Predictor. 
		Given a sequence of features, the model learns to predict the visual feature of the following frame, either in the future or, if the sequence is in reverse order, in the past. The output of the encoder, $\mathbf{r}_t$, corresponds to the visual context at time-step $t$.}
	\label{fig:seq2seq}
\end{figure}

The sequential and relational nature of lifelogging photo-streams allows us to train the weights of an LSTM-based aggregation network without ground truth annotations. To obtain the weights of our Visual Context Predictor, we train an encoder-decoder architecture that, given a sequence of visual feature vectors, learns to predict the subsequent sequence, as shown in Fig.~\ref{fig:seq2seq}. Since LTR video frames are visually highly different from adjacent ones, the model will learn the general context of the event at the same time as the estimation of the visual feature of the upcoming frame.

The auto-encoder is defined as 
\begin{equation}
\label{eq:VCP}
\begin{split}
\mathbf{r_t} & = \mathbf{h_{t, encoder}(x_t)} \\
\mathbf{\hat{x}_{t+1}} & = \mathbf{h_{t, decoder}(r_t)} \, ,
\end{split}
\end{equation}
\noindent where $\mathbf{x_t}$ is the deep learned visual feature (\cf Section~\ref{sec:data_setup}) of frame~$t$, $\mathbf{r_t}$ is the predicted visual context at time $t$, and $\mathbf{h_{t, encoder}}$ and $\mathbf{h_{t, encoder}}$ correspond to the models trained to encode and decode the visual feature, respectively. The objective function of the learning process is to minimize the mean squared error of the prediction, \ie $mse(\mathbf{x_{t}}, \mathbf{\hat{x}_{t}})$.

VCP shares architecture and weights with the encoding model presented above, and is able to encode the visual context of lifelog image sequences both feed forward and backwards, \ie in reverse time order. The chosen architecture for VCP (\ie the encoder) is a single LSTM layer of $1024$ neurons. The hidden state is then passed to the decoder, which has a corresponding LSTM layer. The pre-trained model will be made available upon publication.

\begin{algorithm}[t!]
	\caption{\small Overview of Contextual Event Segmentation}
	\label{al:Alg}
	\small
	\emph{$\triangleright$ Get past and future context from the Visual Context Predictor:}\\
	$\mathbf{rf}(t-1) \leftarrow $ predicted from  $[\mathbf{x}_k]_{\forall ~ 0 ~\leq ~k ~< t}$\\
	
	$\mathbf{rp}(t+1) \leftarrow $ predicted from $[\mathbf{x}_k]_{\forall ~ \mathbf{len}\,[\mathbf{x}] ~\geq ~k ~> t}$\\
	\hrulefill \\
	\emph{$\triangleright$ Detect boundary candidates:}\\
	$pred(t) = cos\_dist(\mathbf{rf}(t-1), \mathbf{rp}(t+1)) $ \\
	$b = \{t \,  | \, (\dfrac{\delta pred}{\delta t} = 0 )\}$ \\
	\emph{$\triangleright$ Remove noisy candidates:}\\
	$b = \{b_k  \, | \,  pred(b_k) \leq average(pred(b))\}$ \\
\end{algorithm}

\subsection{Boundary detector}
Given a frame $\mathbf{x_t}$, two different context predictions can be obtained from VCP. The first, the future context $\mathbf{rf_t}$ including the sequence of frames from the past ($\mathbf{x_{k | 0 \leq k < t}}$). The second, the past context $\mathbf{rp_t}$ including the frames in the future ($\mathbf{x_{k | T \geq k > t}}$), where $T$ is the total length of the lifelog. Thus, at each time-step $t$, the future context given the past will be $\mathbf{rf_{t-1}}$, and the past context given the future $\mathbf{rp_{t+1}}$. Note that the frame $\mathbf{x_t}$ is not seen when predicting the future and past context at time $t$ to avoid overlapping inputs in the prediction.

An event boundary will delimit sequences with very different visual context. Hence, the boundary prediction function is defined
\begin{equation}
\label{eq:pred}
pred(t) = d(\mathbf{rf_{t-1}}, \mathbf{rp_{t+1}}),
\end{equation}
\noindent where $d(\cdot, \cdot)$ is the cosinus distance.

The larger the distance between the two predicted visual contexts, the more likely the upcoming frame will correspond to an event boundary. Since the visual context will change gradually within the vicinity of a boundary, boundary candidates are assigned to the local maxima. Local maximums will also be found for very slight changes in the visual context. Therefore, only the candidates whose prediction value is over the average candidate values are kept as final event boundaries.

\section{R3 dataset}
A large-scale FPV dataset is needed to train the Visual Context Predictor. Such dataset must consist of continuous LTR streams of images spanning at least a few hours, without the need for any annotation. However, the size of the publicly available LTR datasets is very limited: $170$ days in CLEF~\cite{LifeLogTask17_CLEF} and NTCIR~\cite{NTCIR_db}, and $66$ in EDUB-Seg~\cite{dimiccoli2017sr} and EDUB-SegDesc~\cite{bolanos2018egocentric}, spanning a total of $2,700$ hours and $261,845$ images. We can also resort to other popular HTR FPV video datasets such as the First Person Social Interaction Dataset~\cite{FathiDisney}, Huji EgoSet~\cite{PolegSeg}, and UTEgocentric \cite{LeeDisco12}, that cover $28$, $15$ and $16$ hours, respectively. Down-sampled at $2fpm$, the accumulated length of these datasets is under $10,000$ images. This amount of information results insufficient to train efficient deep learning models.

\begin{figure}[t!]
	\centering
	\begin{subfigure}[b]{.485\linewidth}
		\centering
		\includegraphics[width=\linewidth]{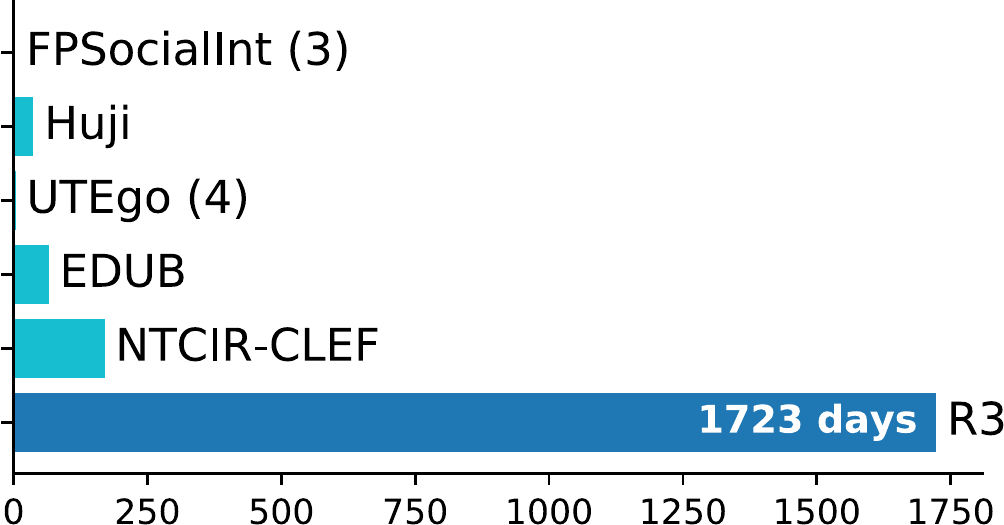}
		\caption{\label{fig:db:days} Number of days}
	\end{subfigure} \,	
	\begin{subfigure}[b]{.485\linewidth}
		\centering
		\includegraphics[width=\linewidth]{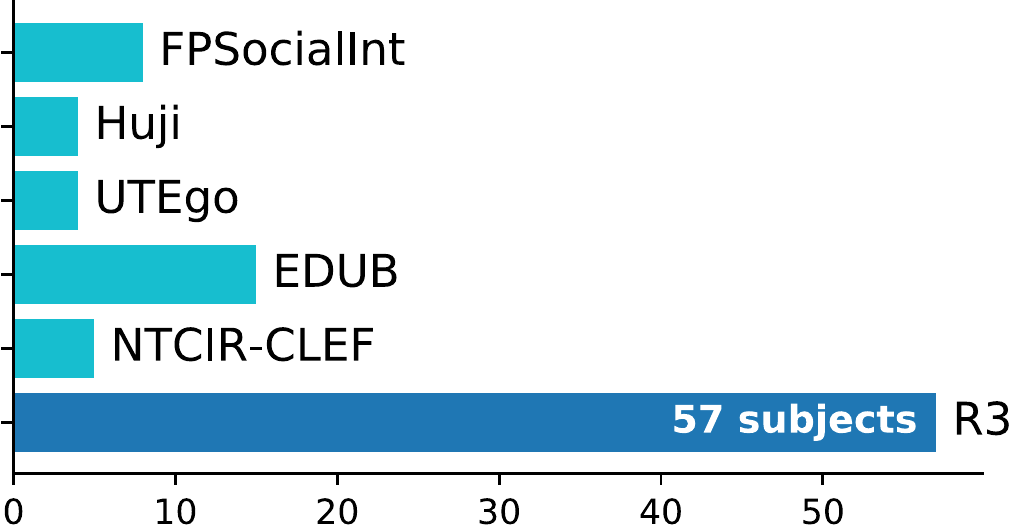}
		\caption{\label{fig:db:subjects} Number of subjects}
	\end{subfigure}	
	\vspace{3pt}
	\begin{subfigure}[b]{1\linewidth}
		\centering
		\includegraphics[width=\linewidth]{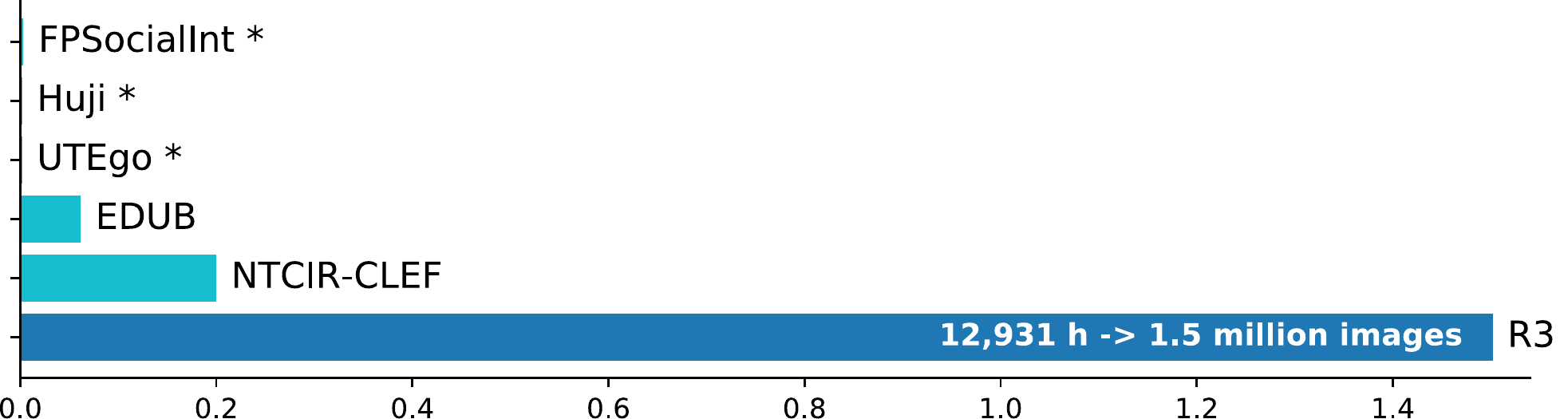}
		\caption{\label{fig:db:images} Total amount of images (in millions).}
		\vspace{-2ex}
	\end{subfigure}
	\vspace{-2ex}
	\caption{Comparison of \emph{R3} with respect to other popular FPV datasets. * HTR datasets are down-sampled at $2fpm$.}
	\label{fig:datasetr}
\end{figure}

In this work we introduce \emph{R3}, a large scale lifelogging image dataset captured by $57$ users during $1,723$ days for a total of almost $13,000$ hours, resulting in over $1.5$ million images. A comparison of the size of \emph{R3} with respect to the other mentioned datasets is presented in Fig.~\ref{fig:datasetr}. The users volunteered to capture their daily lives as part of a memory-enhancement user study. They were asked to put on the wearable camera for most of their day during a whole month, and were free to withdraw from the study if they felt that wearing it was disrupting their routines. The volunteers are mostly seniors older than $50$ years old, and span a wide range of occupations and lifestyles. To protect their privacy, only the extracted visual features will be released.

\section{Experiments}
\subsection{Data setup}
\label{sec:data_setup}
The output of the pre-pooling layer of InceptionV3~\cite{szegedy2016rethinking} is used to describe the frames in the lifelog. We use the available lifelogging video data from R3, CLEF, NTCIR, and EDUB-Seg to train the VCP model and test our CES framework. EDUB-SegDesc~\cite{bolanos2018egocentric} is reserved as validation for further supervised pruning of the prediction obtained from CES. 

The datasets are used as follows:

\textit{Training of the VCP model:} $75\%$ of R3 is used as training set for the Visual Context Predictor model. To ensure that the model is not biased toward this dataset, a $20\%$ of both CLEF~\cite{LifeLogTask17_CLEF} and NTCIR~\cite{NTCIR_db} is also included in the training set. This joined set adds up to $1,207,483$ images. A separate $5\%$ of R3 is used to validate the different configurations and select the best hyperparameters. 

\textit{Testing set for the VCP model:} the remaining $20\%$ of R3, and $80\%$ of CLEF and NTCIR is kept as test to confirm that VCP is not overfitted toward R3 (\cf~\ref{tab:mse}).

 \textit{Testing of the CES framework:}  the semantic features for $12$ of the lifelogs in EDUB-Seg~\cite{dimiccoli2017sr} have been made available to us. We compare our method to the baselines in two overlapping sets: these $12$ lifelogs and the full $20$ lifelogs in the dataset.


\subsection{Training methodology}

We explore several architectures and training parameters for the Visual Context Predictor model. Regarding the architecture, we can modify the number of neurons in the encoding LSTM layer, the number of frames seen before starting the future prediction ($N$), the amount of frames the decoder needs to predict ($M$), and whether the prediction will be conditional or not, \ie whether the model gets further inputs past frame $N$. 
We investigate architectures between $256$ and $1024$ neurons, values of $N=M$ between $10$ and $100$, and the same range of $M$ for $N=1$. 

Concerning the training parameters, the loss is defined as the mean squared error of the prediction $\mathbf{\hat{x}_t}$, and RMSProp without decay is used as optimizer. The learning rate is randomly set in the range $[.0001, .001]$, and is reduced by half after every $4$ epochs without significant improvement in the validation loss. Different batch sizes are used, between $250$ and $1000$ sequences at a time.

The best configuration is found through a gridsearch on all the different parameters. We find that the best prediction performance (smaller validation loss) is achieved with $1024$ neurons on a conditional architecture. The number of frames seen before starting the future prediction is set to $N=10$, equal to the number of frames to predict ($M=10$). We observe that training with longer sequences does not improve significantly the model performance (\cf Table~\ref{tab:mse}), while making the training slower. At test time, one single frame ($N=1$) is given to start the prediction of the whole day ($M=length(lifelog)-1$).

\paragraph{Other implementation details.}
We also analyze the possibility of fine-tuning the boundary prediction with supervised learning. For that purpose, we train an SVM with samples from a held-out validation set (EDUB-SegDesc~\cite{bolanos2018egocentric}). The SVM evaluates the boundary likeliness from cluster consistency indicators. In particular, two clusters are defined at opposite sides of the candidate boundary, containing the $15$ frames that precede or follow it. The indicators used are the correlation between the two clusters, the compactness of each of them and their union, and the BetaCV and Normalized Cut scores~\cite{zaki2014data}.

\begin{figure}[t]
	\centering
	\includegraphics[width=\linewidth]{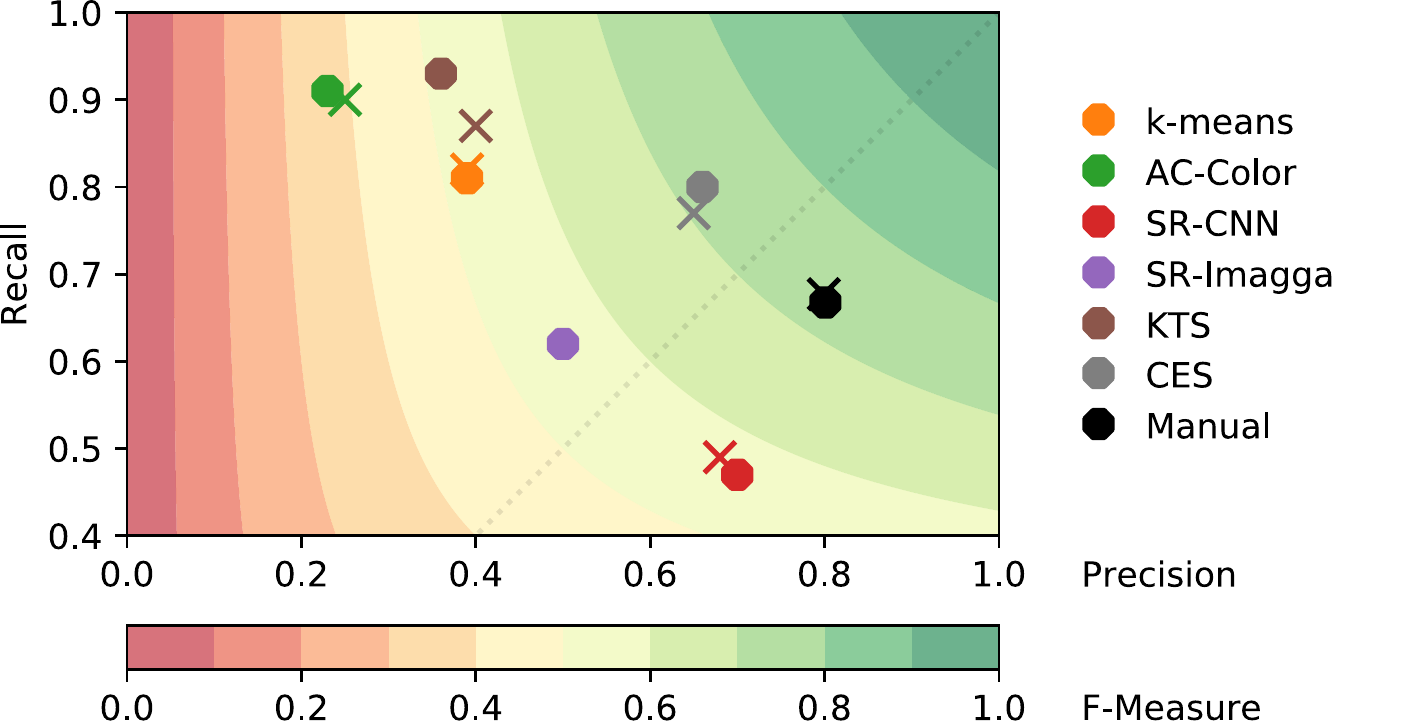}
	\vspace*{-2pt}
	\caption{Precision-Recall curve for the tested unsupervised event segmentation algorithms. The corresponding F-measure score is shown in the color space. Results on the smaller set EDUB-Seg12 are represented with an octagon, whereas the larger set EDUB-Seg20 is depicted with an x. (Best viewed in color.)}
	\label{fig:PRcurve}
\end{figure}

\subsection{Evaluation methodology}
Following the literature, we report the averaged f-measure, precision and recall for the tested models (Table~\ref{tab:results}). 
For our evaluation, a detected boundary is considered a true positive if there is an element in the ground truth within a distance of tolerance, and the ground truth element is not already matched to any other detected boundary. Analogously, all elements in the ground truth for which no detected boundary is found within the tolerance are considered false negatives. This tolerance is set to $5$ frames.

We compare the performance of the following baselines on the publicly available \emph{EDUB-Seg} dataset~\cite{dimiccoli2017sr}:
\begin{itemize}
\item{\textbf{Smoothed K-Means}:} the lifelog is clustered into events using k-means with a fixed $k=30$. The clustering is then smoothed by assigning each frame to the most common cluster within a window. This operation is done iteratively until no more changes occur. As a result, some clusters may disappear.
\item{\textbf{AC-Color}:} Agglomerative Clustering on the color feature of the frames, as done in~\cite{LeePred15}.
\item{\textbf{SR-Clustering}:} Semantic Regularized Clustering as de-\break scribed in~\cite{dimiccoli2017sr}, using only visual features (\emph{CNN}), and also semantic cues (\emph{Imagga}).
\item{\textbf{KTS}:} Kernel Temporal Segmentation as described in~\cite{PotapovSum}.
\end{itemize}

\paragraph{Bias in the Ground Truth}
Since segmenting lifelogs into events can be a very subjective task, the curators of EDUB-Seg provide in ~\cite{dimiccoli2017sr} an extensive analysis on the uniformity among the ground truth annotated by different subjects. They conclude that visual lifelog event segmentation can be objectively evaluated, since different people (which are not the camera wearer) tend to segment the lifelogs consistently. For the purpose of our evaluation, we select the ground truth from the first annotator. We use the other annotations as a baseline. For the lifelogs that only included one annotation, we asked independent subjects to annotate the events, so that we would have at least two sets of annotations for each lifelog. We therefore report the performance of the manual annotations as an upper reference in Table~\ref{tab:results:detailed}.

\paragraph{Other implementation details.}
To find the local maximums in the prediction signal of CES, as well as smoothing the K-Means clustering, a window of size $5$ is chosen, so that it is consistent with the ground truth tolerance.

\begin{table}[t!]
	\centering
	\setlength\extrarowheight{.2ex}
	\setlength{\tabcolsep}{3pt}
	\begin{tabular}[width=\textwidth]{@{}l|ccc|ccc}
		\multicolumn{1}{@{}c@{}}{}& \multicolumn{3}{@{}c@{}}{EDUB-Seg12}  &  \multicolumn{3}{@{}c@{}}{EDUB-Seg20}\\
		\multicolumn{1}{@{}c@{}|}{method} & F1  & Prec. & Rec. & F1 & Prec. & Rec. \\ \hline
		K-Means smoothed & 0.51 & 0.39 & 0.81 & 0.51 & 0.39 & 0.82\\
		AC-Color~\cite{LeePred15} &  0.36 & 0.23 & 0.91 & 0.38 & 0.25 & 0.90 \\
		SR-ClusteringCNN~\cite{dimiccoli2017sr}& 0.50 & 0.70 & 0.47 & \textbf{0.53} & 0.68 & 0.49\\
		SR-ClusteringImagga~\cite{dimiccoli2017sr} & \textbf{0.53} & 0.50 & 0.62 & &-&\\
		KTS~\cite{PotapovSum}& 0.50 & 0.36 & 0.93 & \textbf{0.53} & 0.40 & 0.87\\[.7ex] \hline 
		\textbf{CES (with VCP)} & \textbf{\textcolor{darkergreen}{0.70}} & 0.66 & 0.80 & \textbf{\textcolor{darkergreen}{0.69}} & 0.66 & 0.77\\ 
	\end{tabular}
	\caption{Comparison to the state of the art. Averaged results (F-measure, Precision and Recall) on the subset of $12$ lifelogs with available semantic tags and on the full EDUB-Seg.}
	\label{tab:results}
\end{table}

\begin{figure*}[t!]
	\centering
	\includegraphics[width=\linewidth]{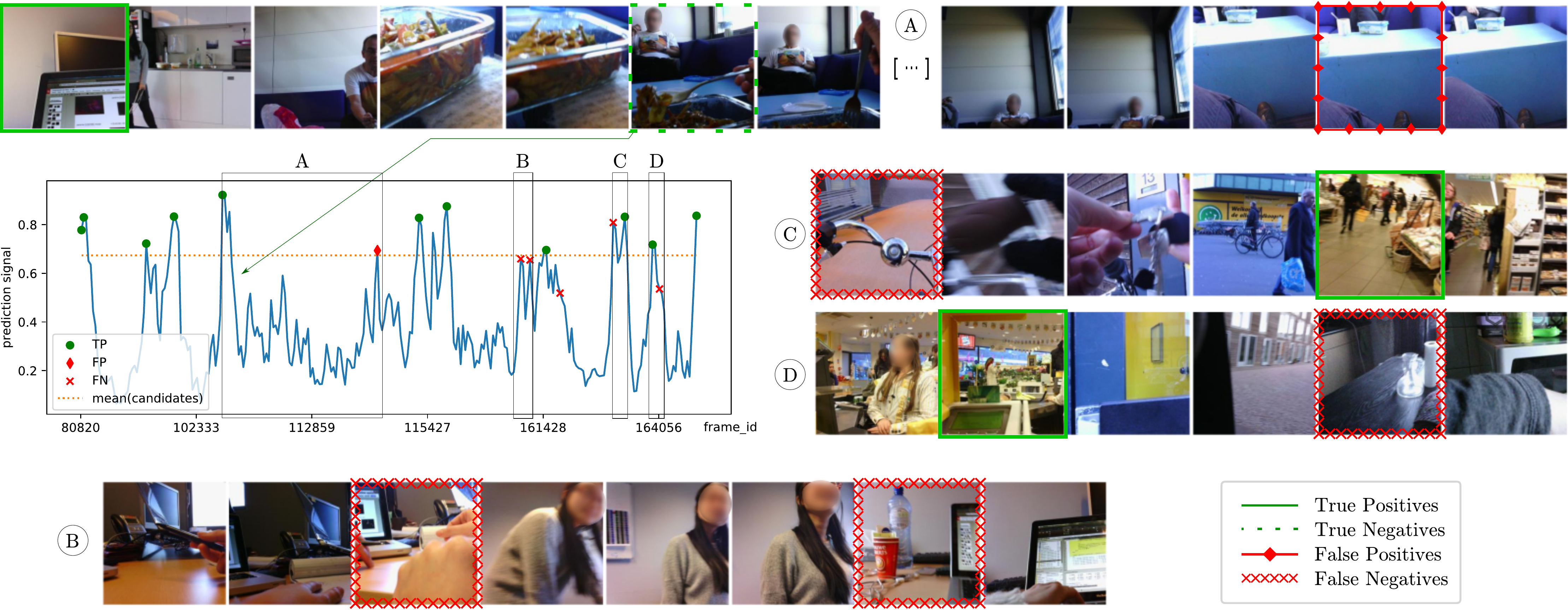}
	\caption{Qualitative example for one of the tested lifelogs. We highlight those frames which are false positives or false negatives in the baselines. We can observe that, unlike the baselines, CES is able to ignore occasional occlusions as long as the different points of view span less frames than CES' memory span (A). It is also capable of detecting boundaries that separate heterogeneous events such as riding a bike on the street and shopping at the supermarket (C, D). Most of the boundaries not detected by CES correspond to events that take place within the same physical space (B) and short transitions (C, D), \eg parking the bike. (Best viewed in color.)}
	\label{fig:res}
\end{figure*}


\subsection{Results}
Table~\ref{tab:results} presents the results of CES and the baselines in EDUB-Seg, and a smaller subset (which includes the semantic features needed for \textit{SR-ClusteringImagga}). The position of each method in the Precision-Recall curve is shown in Fig.~\ref{fig:PRcurve}. While most methods fall within the mid-range performance in terms of f-measure, CES stands out of the baselines, improving their performance by over $15\%$, and positioning itself on the upper range of the absolute spectrum. The performance of CES is even competitive with that of the manual annotations. 

We show in Fig.~\ref{fig:res} the performance of CES applied to one of the tested lifelogs. We can observe that most elements in the ground truth fall on the spikes of the prediction signal, or very close to them. This confirms the suitability of using the predicted contexts as a boundary cue.

\begin{figure*}[t]
\centering
\begin{subfigure}[b]{\linewidth}
            \centering           
\cfbox{lightgray}{\includegraphics[width=.09\linewidth]{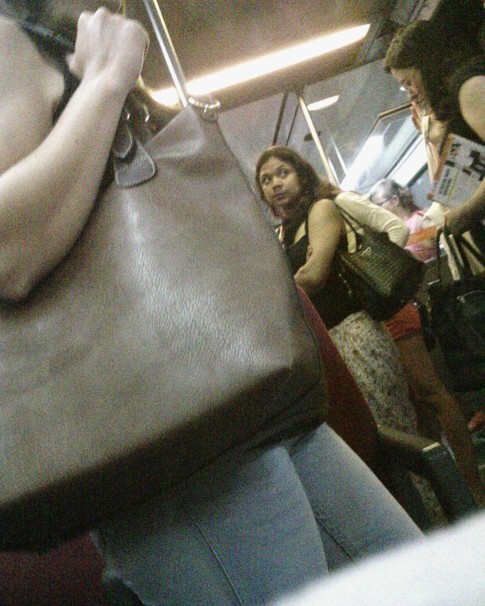}
\includegraphics[width=.09\linewidth]{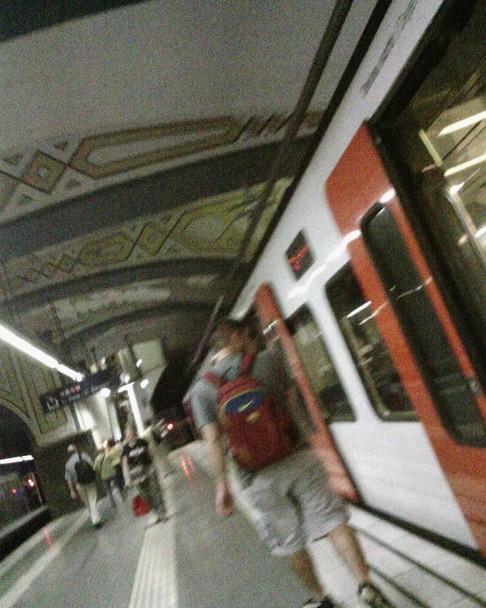}
\includegraphics[width=.09\linewidth]{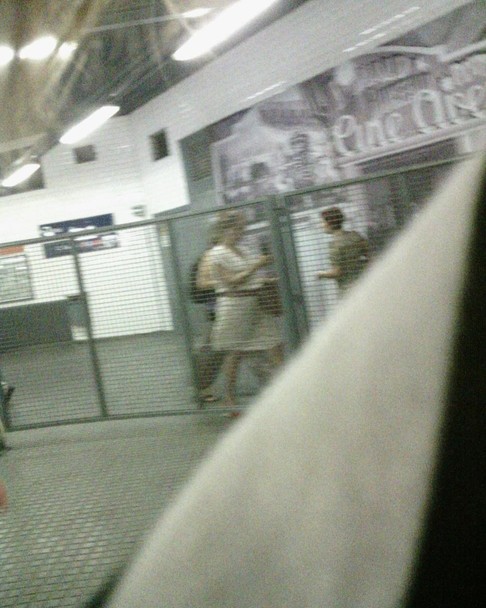}
\includegraphics[width=.09\linewidth]{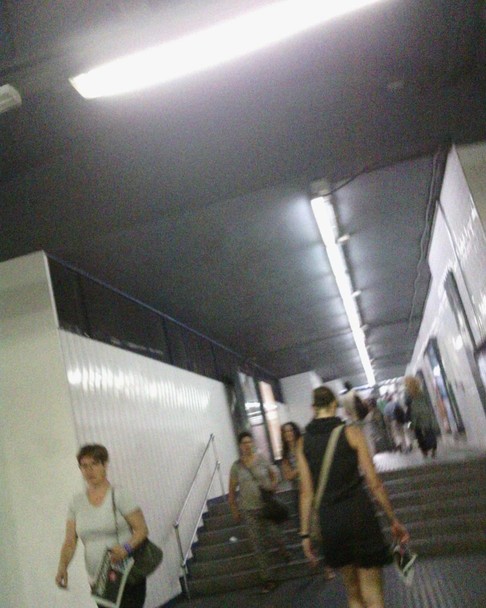}
\includegraphics[width=.09\linewidth]{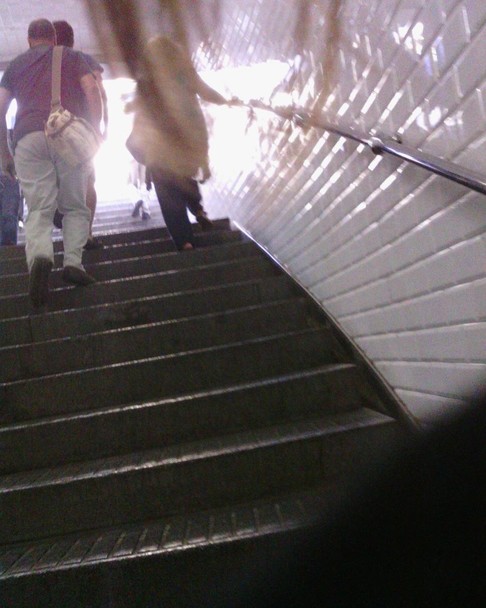}}
\, \raisebox{.5em}{\textbf{\textcolor{darkgreen}{TP}}} \,
\cfbox{lightorange}{\includegraphics[width=.09\linewidth]{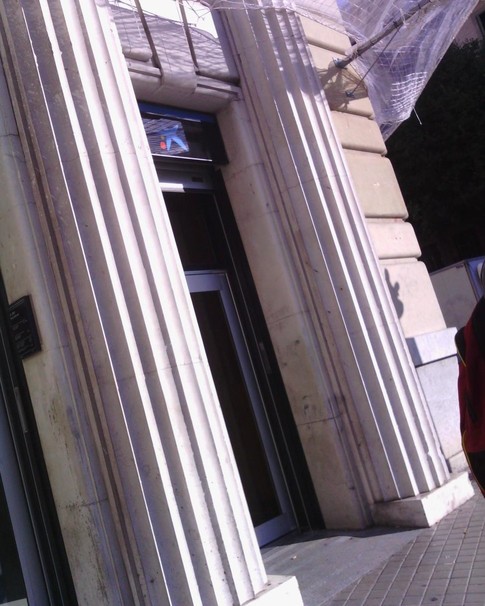}
\includegraphics[width=.09\linewidth]{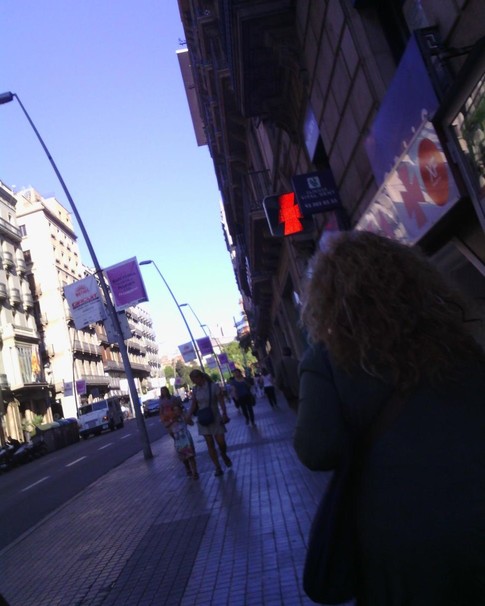}
\includegraphics[width=.09\linewidth]{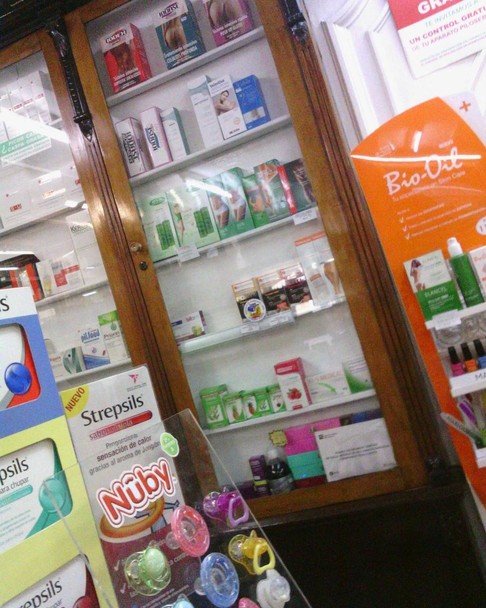}
\includegraphics[width=.09\linewidth]{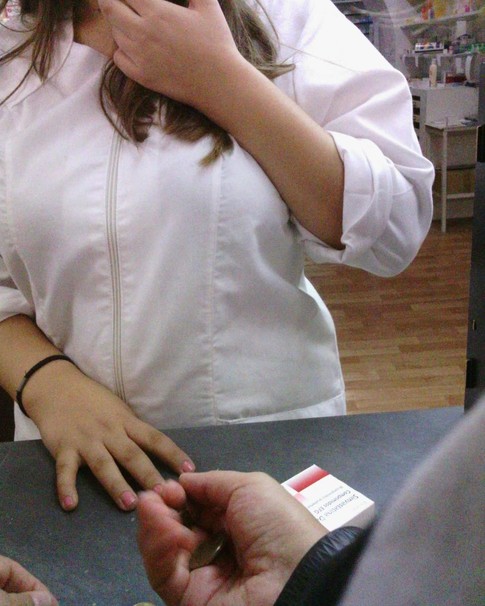}
\includegraphics[width=.09\linewidth]{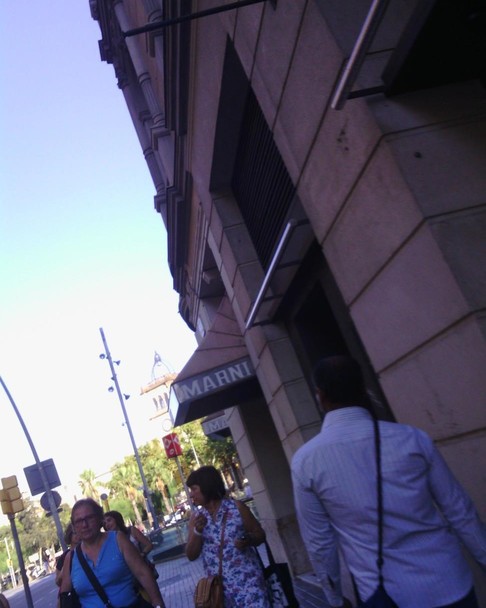}}

    \vspace{-.4em}
     \caption{\label{fig:res:tp} True Positives: CES can model public transportation events, as well as street walking.}
     \vspace{.5em}
    \end{subfigure}
\begin{subfigure}[b]{\linewidth}
            \centering
\cfbox{lightgray}{\includegraphics[width=.09\linewidth]{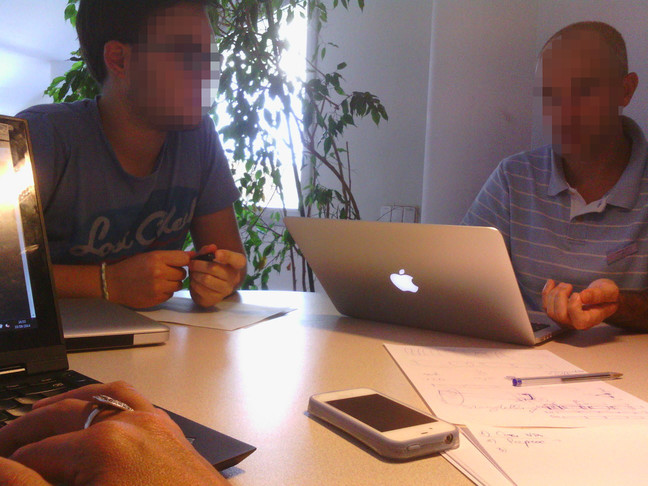}
\includegraphics[width=.09\linewidth]{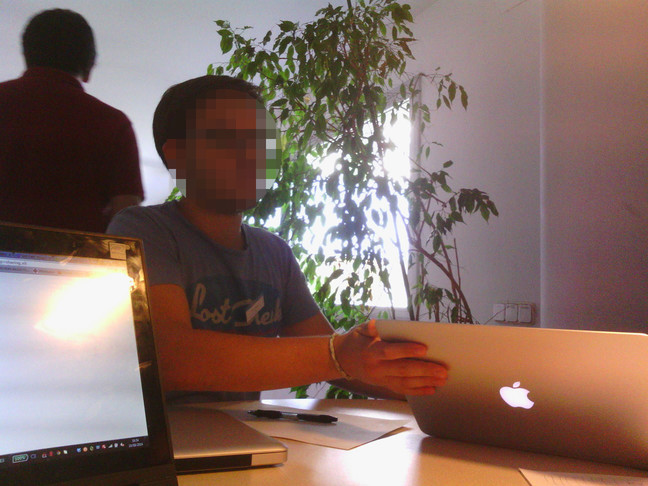}
\includegraphics[width=.09\linewidth]{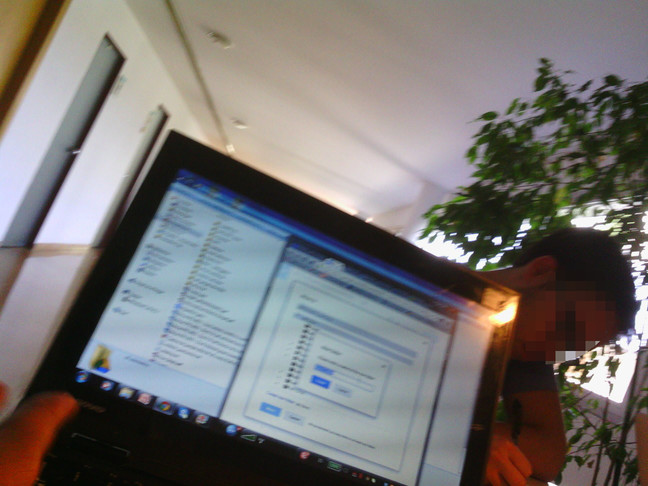}
\includegraphics[width=.09\linewidth]{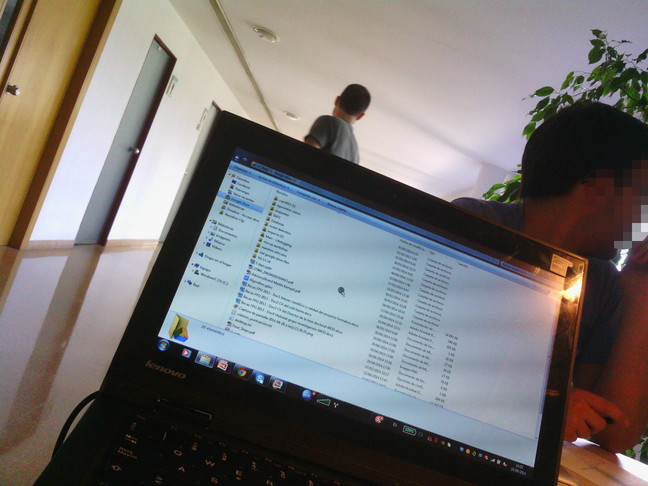}
\includegraphics[width=.09\linewidth]{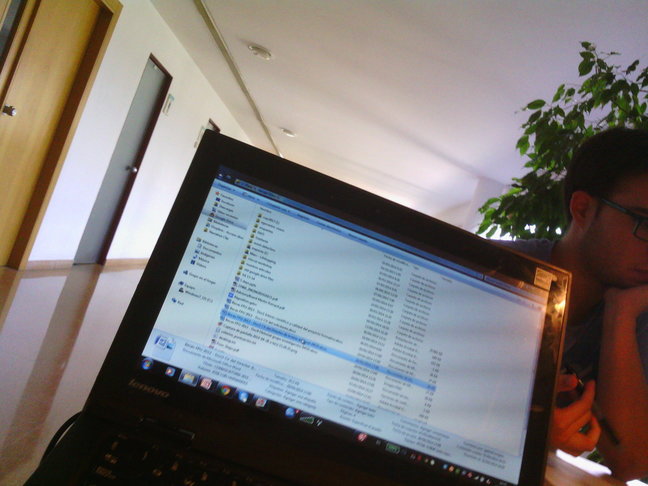}
\, \raisebox{.5em}{\textcolor{darkgreen}{\textbf{TN}}} \,
\includegraphics[width=.09\linewidth]{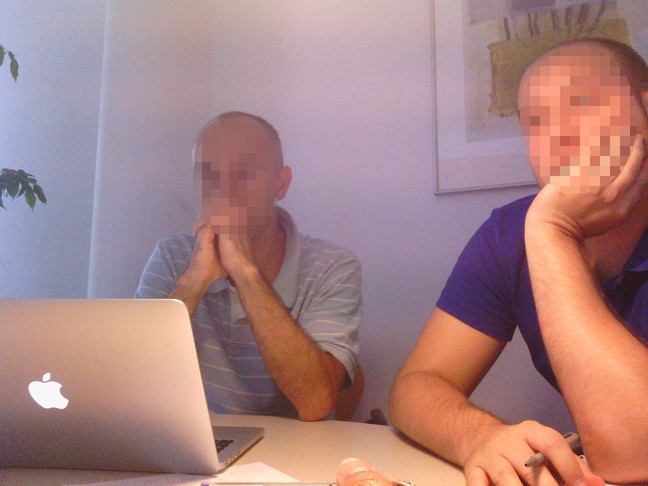}
\includegraphics[width=.09\linewidth]{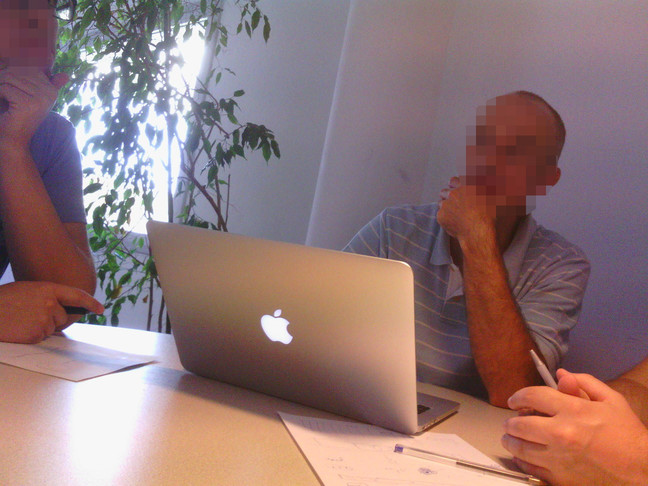}
\includegraphics[width=.09\linewidth]{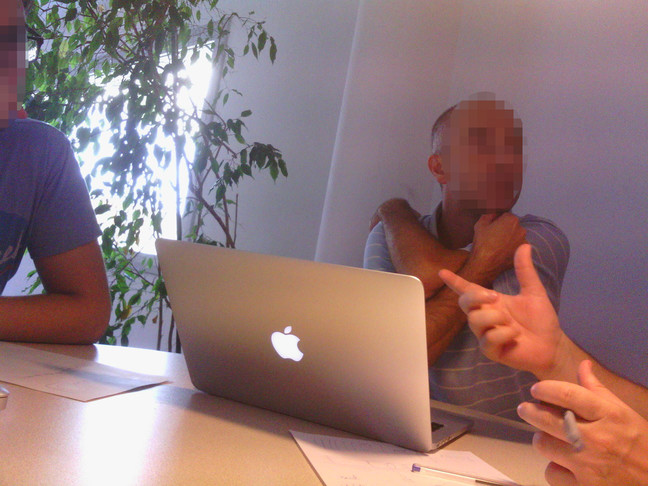}
\includegraphics[width=.09\linewidth]{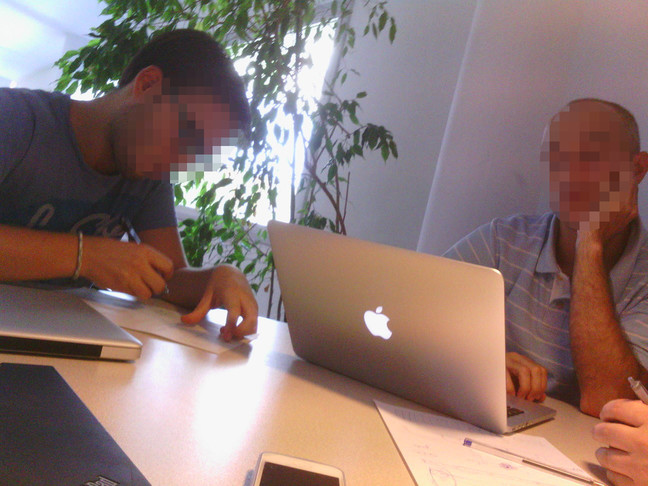}
\includegraphics[width=.09\linewidth]{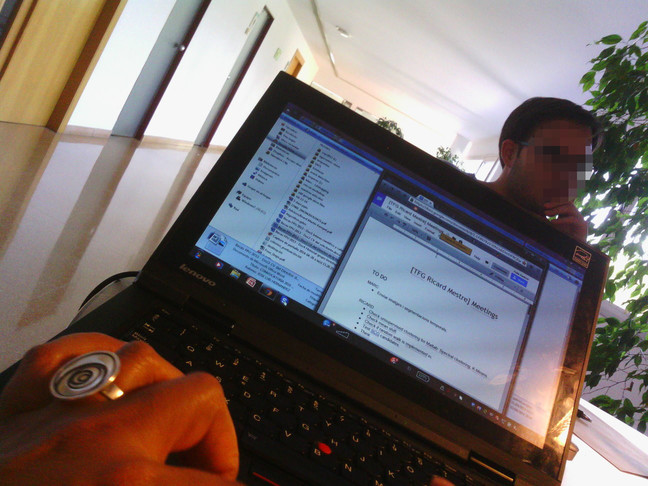}}
\vspace{-.4em}
     \caption{\label{fig:res:tn} True Negatives: CES remembers previously seen context, and is able to match future and past.}
     \vspace{.5em}
    \end{subfigure}
\begin{subfigure}[b]{\linewidth}
            \centering
\cfbox{lightgray}{\includegraphics[width=.09\linewidth]{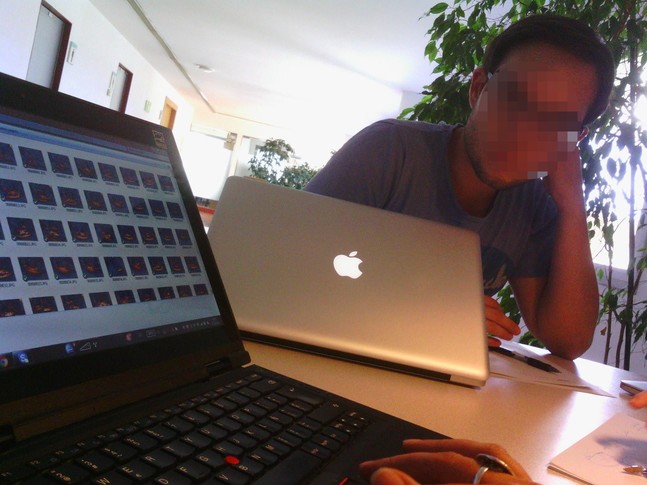}
\includegraphics[width=.09\linewidth]{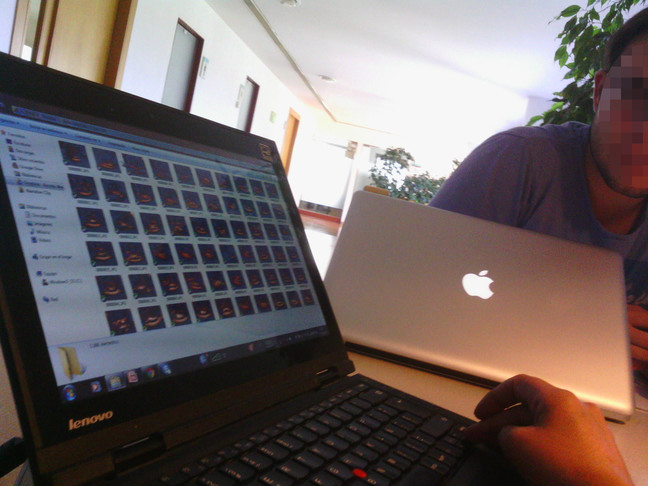}
\includegraphics[width=.09\linewidth]{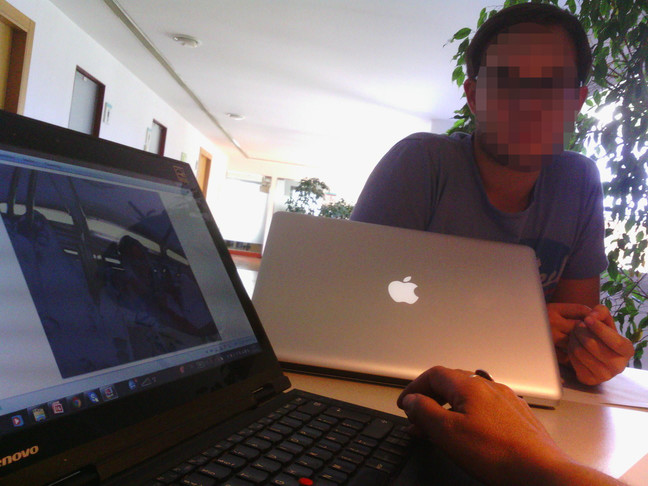}
\includegraphics[width=.09\linewidth]{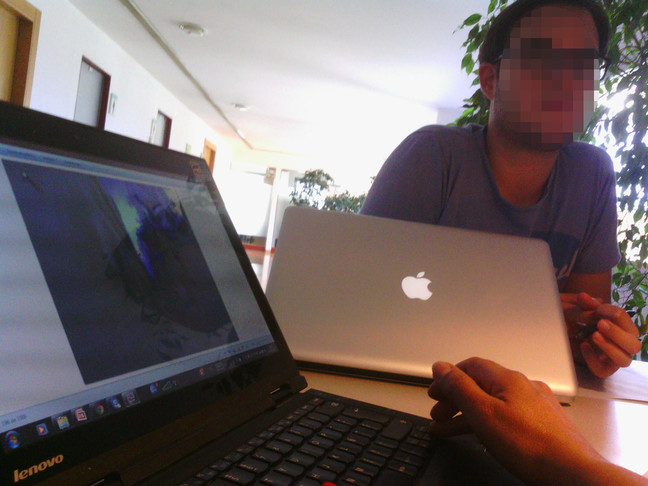}
\includegraphics[width=.09\linewidth]{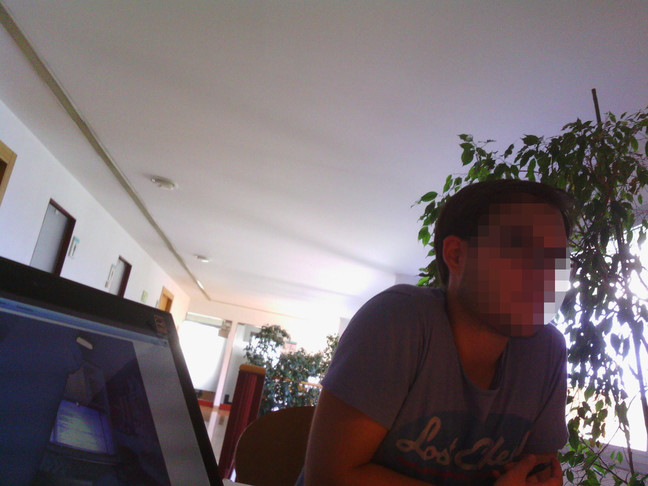}}
\, \raisebox{.5em}{\textcolor{red}{\textbf{FP}}} \, 
\cfbox{lightorange}{\includegraphics[width=.09\linewidth]{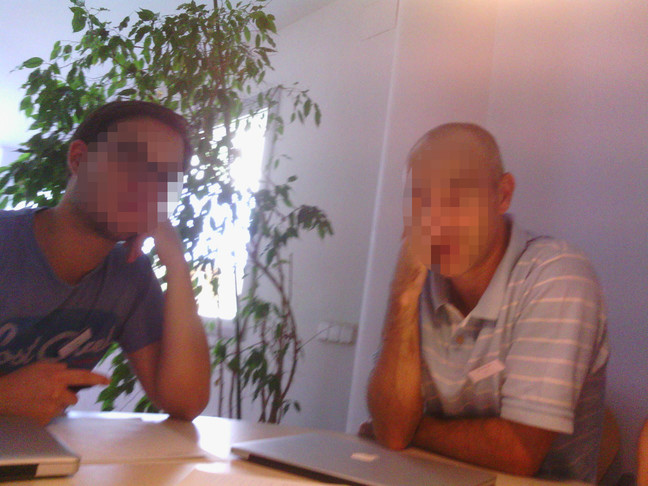}
\includegraphics[width=.09\linewidth]{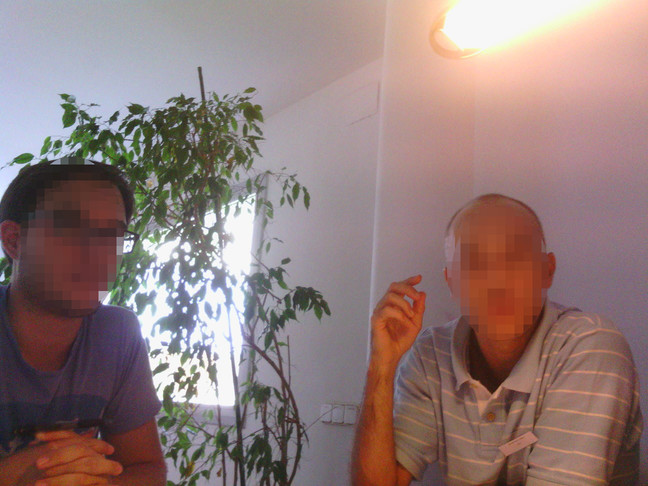}
\includegraphics[width=.09\linewidth]{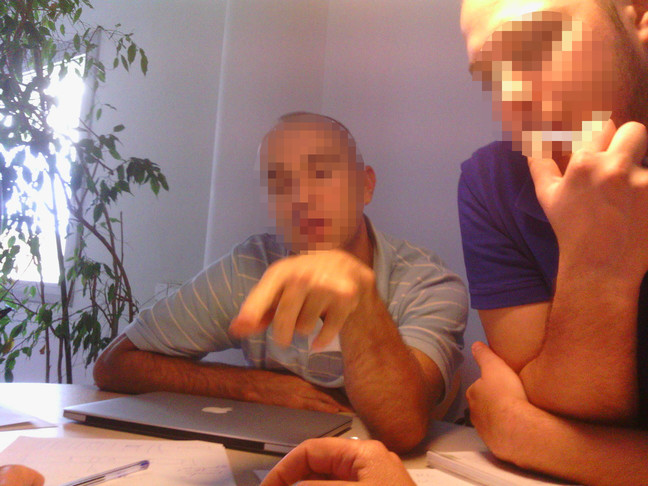}
\includegraphics[width=.09\linewidth]{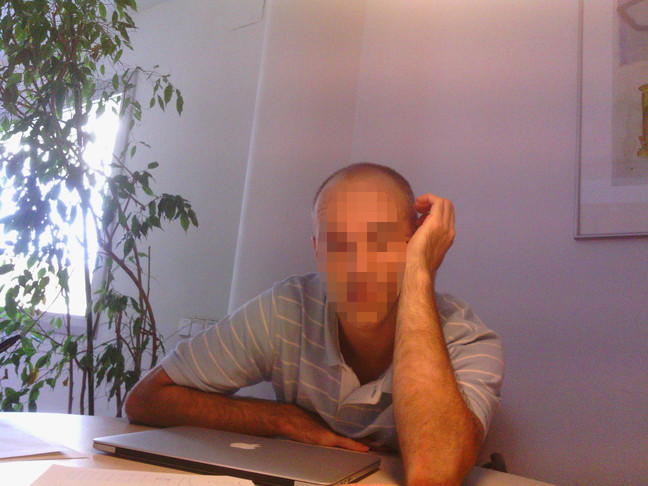}
\includegraphics[width=.09\linewidth]{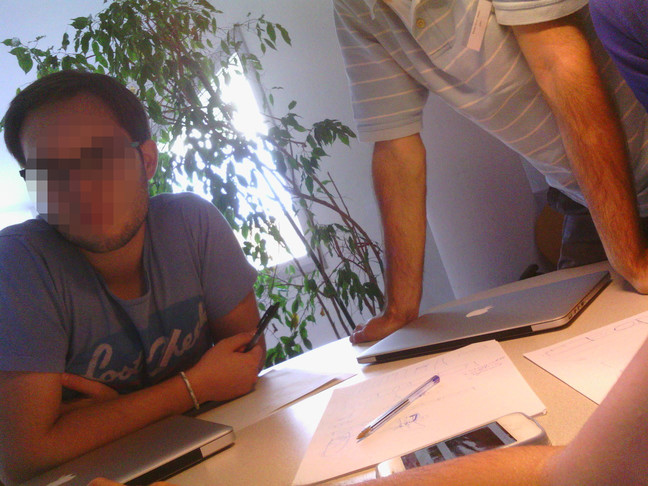}}
	\vspace*{-1.2em}
     \caption{\label{fig:res:fp} False Positives: if the different sight positions span longer than CES' memory span, a false positive will raise.}
     \vspace{.5em}
    \end{subfigure}

 \begin{subfigure}[b]{\linewidth}
             \centering
\cfbox{lightgray}{ \includegraphics[width=.09\linewidth]{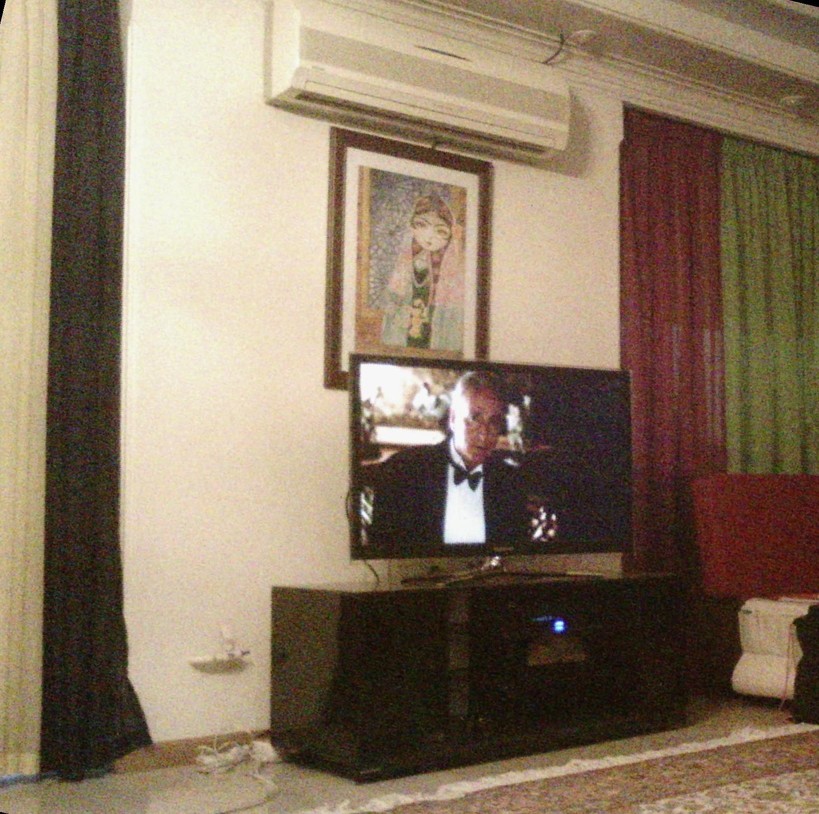}
 \includegraphics[width=.09\linewidth]{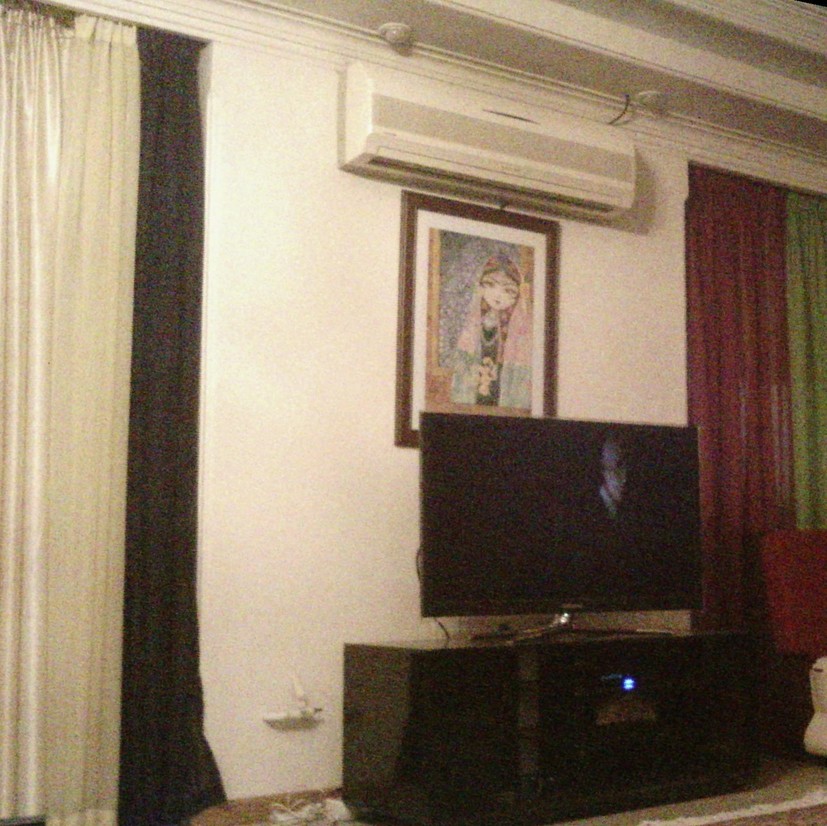}
 \includegraphics[width=.09\linewidth]{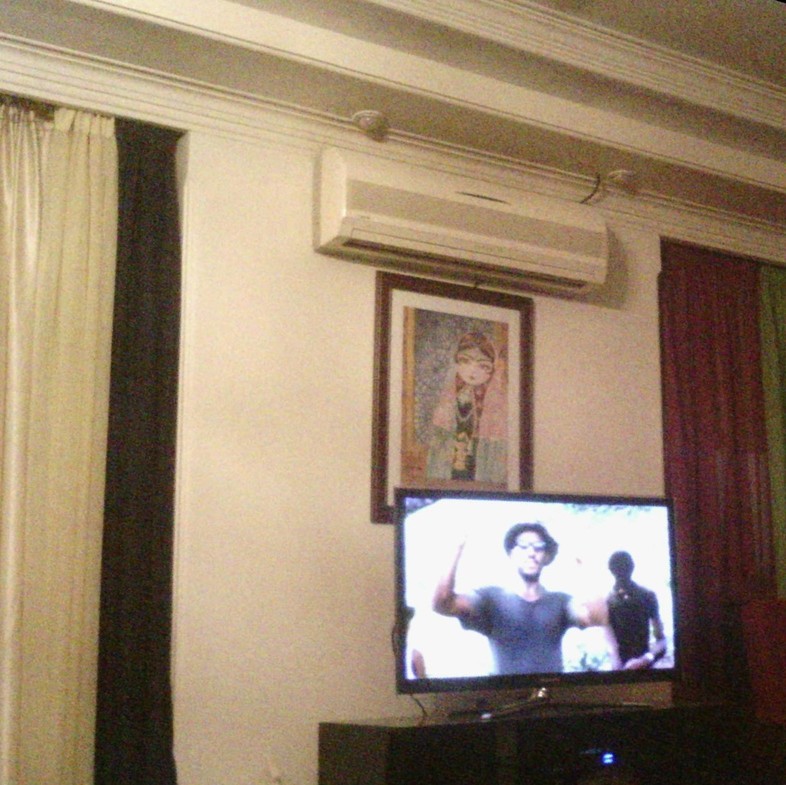}
 \includegraphics[width=.09\linewidth]{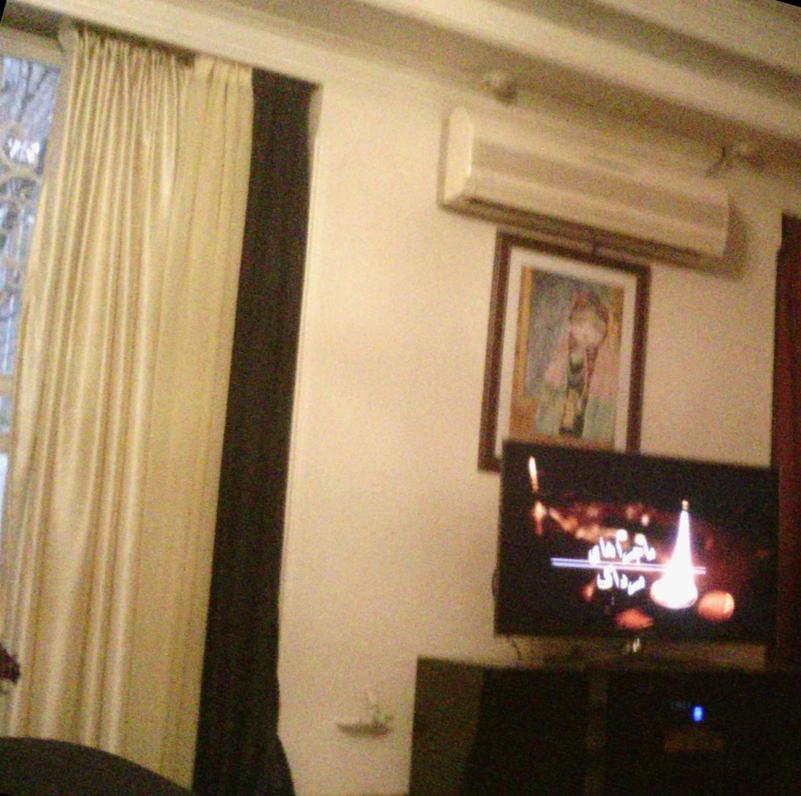}
 \includegraphics[width=.09\linewidth]{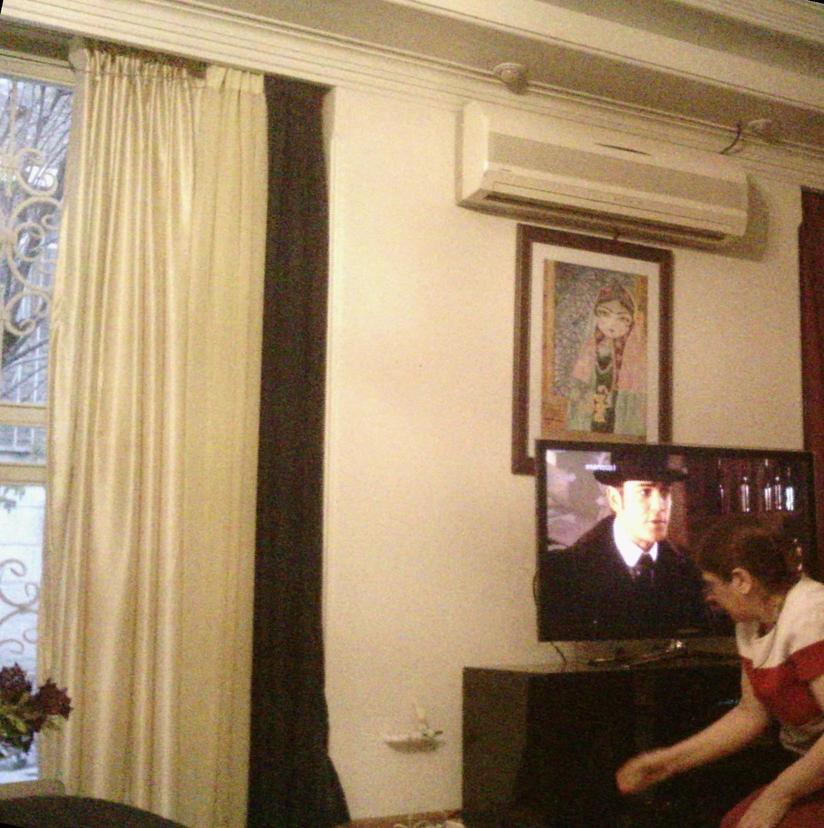}
 \, \raisebox{.5em}{\textcolor{red}{\textbf{FN}}} \,
 \includegraphics[width=.09\linewidth]{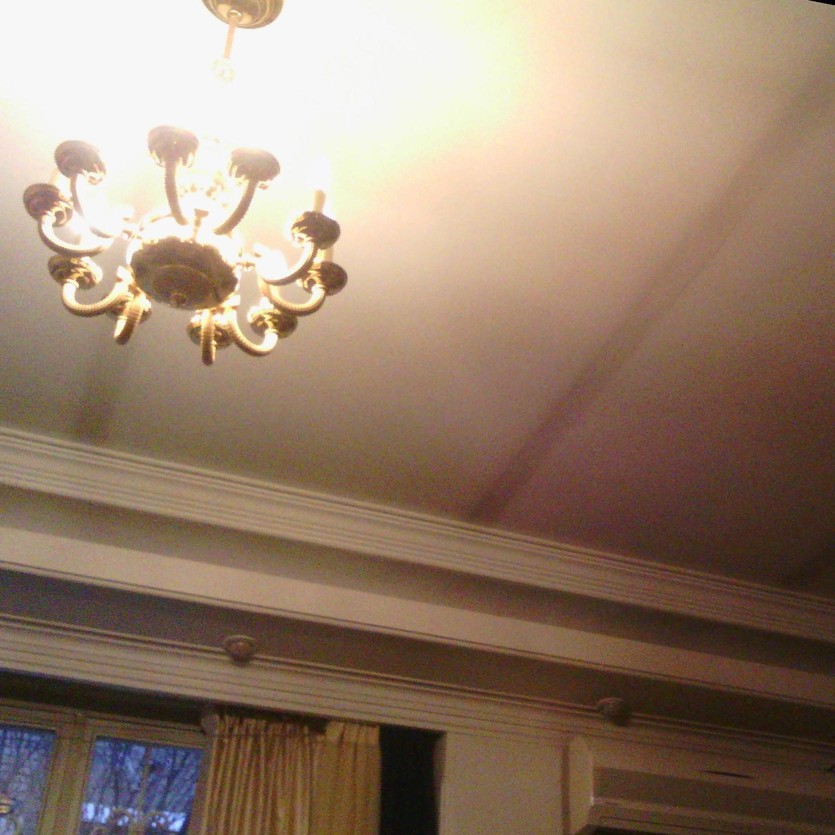}
 \includegraphics[width=.09\linewidth]{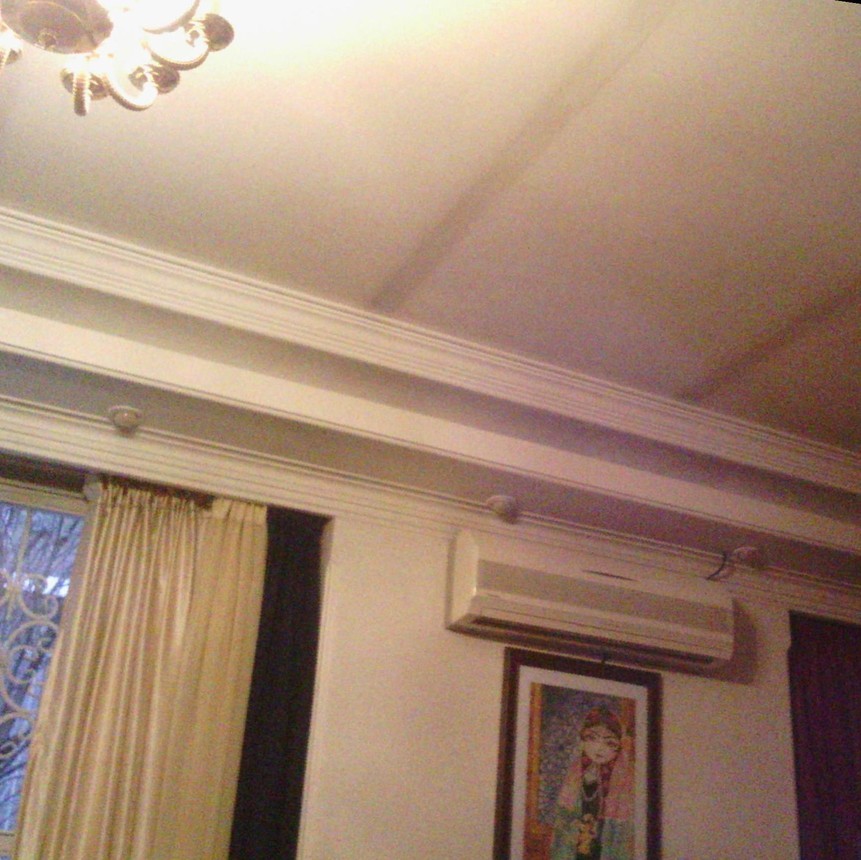}
 \includegraphics[width=.09\linewidth]{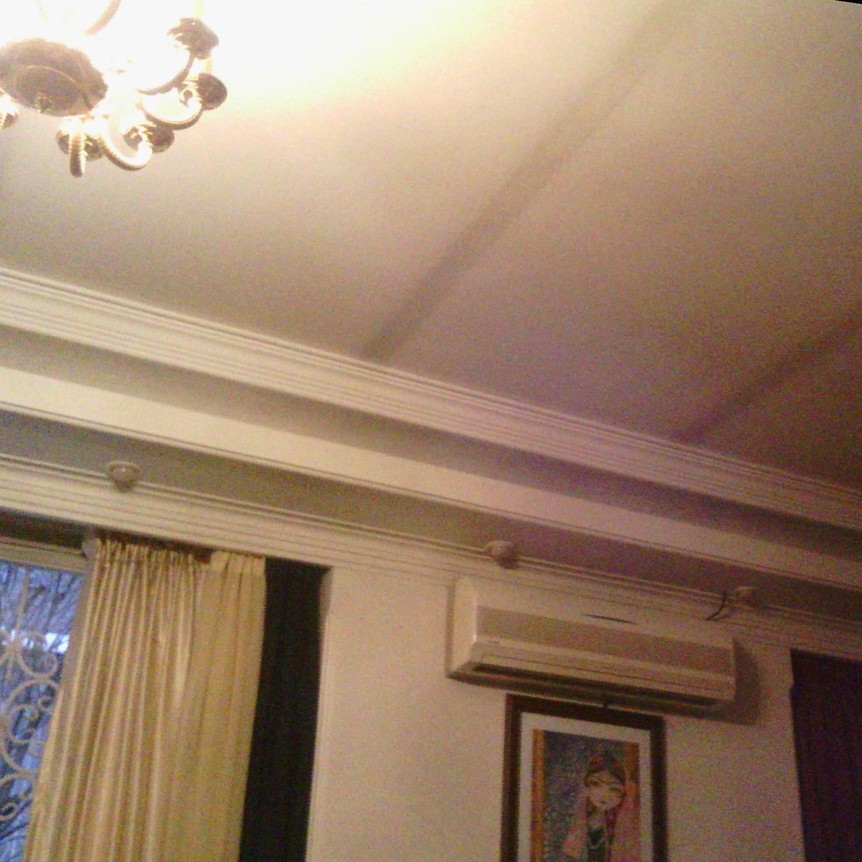}
 \includegraphics[width=.09\linewidth]{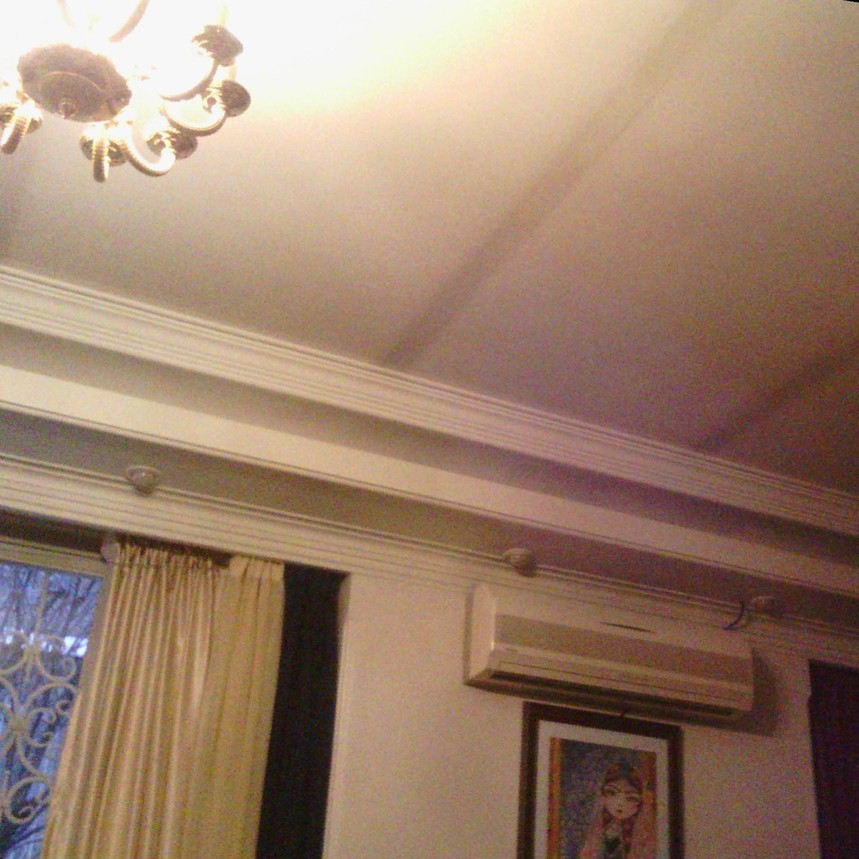}
 \includegraphics[width=.09\linewidth]{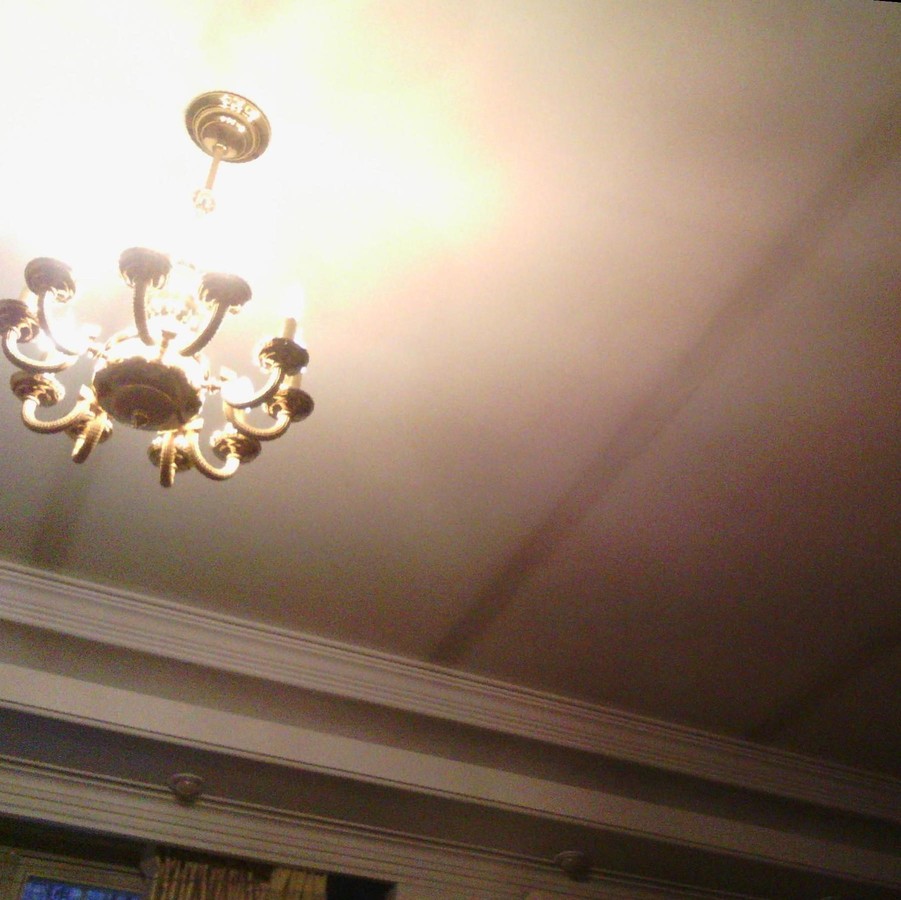}}
 \vspace{-.4em}
      \caption{\label{fig:res:fn} False Negatives: two events taking place in the same location can sometimes be understood as a single one.}
     \end{subfigure}

\vspace*{-2ex}
\caption{Examples of the capacities of CES. The detected events are framed in separate boxes.}
\label{fig:ex2}
\end{figure*}

While the baselines fail at detecting boundaries between heterogeneous events, CES is capable of extracting the underlying context of each event, and discern their disparity (\eg shopping at the supermarket after riding a bike on the street). Moreover, in cases in which the camera wearer orientation changes within a static event (\eg looking back from your food to your colleagues), traditional segmentation methods detect such view change as an event boundary, whereas CES is able to detect the presence of a common visual path. However, if the view change spans longer than CES memory, CES will not be able to contextualize it within the event. An example for such a situation can be seen in Figs.~\ref{fig:res:tn} and ~\ref{fig:res:fp}. We also note that the ability of CES to detect the general context of the visual sequence and track common cues sometimes misleads the prediction. When the ground truth of a boundary falls within the same physical space, or similar contexts, CES does not perceive their differences, and thus does not detect the boundary. Arguably, such boundaries are also difficult to detect by external viewers. This may also occur when short transitions between events are considered events on their own.

\paragraph{Predicting the context vs predicting the actual frame}
One could think that predicting a future frame $\mathbf{\hat{x}_t}$ and comparing it to the actual future frame $\mathbf{x_t}$ should be better than comparing the visual context. We tested this hypothesis, in which
\begin{equation}
\label{eq:pred}
pred(t) = abs(mse(\mathbf{x_t}, \mathbf{\hat{x}_{f\, t}}) - mse(\mathbf{x_t}, \mathbf{\hat{x}_{p \,t}})),
\end{equation}
\noindent where $\mathbf{\hat{x}_{f\, t}}$ is predicted from $\mathbf{x_{k | 0 \leq k < t}}$ and $\mathbf{\hat{x}_{p\, t}}$ from $\mathbf{x_{k | T \geq k > t}}$. The intuition behind this formulation is that a local outlier will be badly predicted both from the future and the past, whereas an event change will provide a good prediction only in one direction.
This theory proves not precise in practice. The generative model embeds noise into the frame descriptor, and, as expected, generates samples closer to the previous (seen) frame than the (unseen) target. As such, using such a noisy signal is detrimental to the final objective. The performance of such method is reported as CES-error in Table~\ref{tab:results:detailed}. 

\paragraph{Informativeness of the Visual Context}
\todo{The experimental results proving that the visual context is an efficient cue for event boundary prediction (\cf Fig.~\ref{fig:res}) suggest that it may also be informative and particular for each speciffic event, such as walking on the street or having lunch with work colleagues. We analyze its informativeness by matching events from different lifelogs, as can be seen in Fig.~\ref{fig:clustering}. We observe that VCP is able to describe similarly the same heterogeneous events from different users in different countries. MORE TO BE FILLED TOMORROW}
To validate the encoding efficiency of VCP and hence the informativeness of the visual context, we have tested CES using two alternative sequence encodings: first, an average of the previous $N=10$ frames (or subsequent in the case of the past prediction); second, a PCA time-dimensionality reduction on the aforesaid set. 
These two variants are reported in Table~\ref{tab:results:detailed} as CES-mean and CES-PCA, respectively.

We observe that the visual context predicted by VCP results much more informative than any of the other contextual encodings. While the averaged encoding obtains a predictive performance similar to the output of our decoder (\cf Table~\ref{tab:mse}), the encoding transformation of VCP is superior as a contextual visual feature. Moreover, unlike PCA, which takes the inputs as a set, VCP takes the inputs as a sequence, and is able to learn a more informative context descriptor.

\iffalse
\begin{table}[b!]
\centering
\setlength\extrarowheight{.2ex}
\setlength{\tabcolsep}{1.5pt}
\begin{tabular}[width=\textwidth]{@{}lccc|ccH|ccH@{}}

trained with N / M : &\multicolumn{3}{@{}c@{}|}{10 / 10}& \multicolumn{3}{@{}c@{}|}{1 / 40} & \multicolumn{3}{@{}c@{}}{1 / 100}\\
\# neurons : &1024 &512 &256 &1024 &512 &256 &1024 &512 &256\\
\hline
mse future prediction: &1.024 &1.030 &1.058 &1.029 &1.030 &1.057 &1.028 &1.029 &1.053\\
mse past prediction: &1.024 &1.029 &1.059 &1.029 &1.030 &1.058 &1.028 &1.029 &1.053\\

\end{tabular}
\caption{Performance at test time 
(mean \emph{mse} amplified $\cdot 10^{2}$, with $N=1$, $M=T-1$ and $T = \mathbf{len}~[\mathbf{x}]$) for different training configurations of VCP. As a reference, note that the performance of the future prediction $\mathbf{\hat{x}}(t) = \mathbf{x}(t-1)$ is of $ 1.58 \cdot 10^{-2}$.}
\label{tab:mse}
\end{table}

\else
\begin{table}[t!]
\centering
\setlength\extrarowheight{.2ex}
\setlength{\tabcolsep}{1.5pt}
\begin{tabular}[width=\textwidth]{@{}l@{}@{}ccc|ccc|@{}c|cc@{}}

trained with N / M : &\multicolumn{3}{@{}c@{}|}{10 / 10}& \multicolumn{2}{@{}c@{}}{1 / 40} & 1/100&& 1/1 & 10/1\\
\# neurons : &256 &512 &1024 &512  &1024  &1024 &&  \multicolumn{2}{@{}c@{}}{mean*} \\
\cline{1-7}\cline{9-10}
mse future pred.:  &1.058&1.030 &\textbf{1.024} &1.03 &1.029 &1.028 && \multirow{2}{*}{1.58}& \multirow{2}{*}{1.054}\\
mse past pred.:  &1.059&1.029 &\textbf{1.024} &1.03 &1.029 &1.028 && &\\

\end{tabular}
\caption{Performance of the auto-encoder's prediction at test time 
(mean \emph{mse} amplified $\cdot 10^{2}$, with $N=1$, $M=T-1$ and $T = \mathbf{len}~[\mathbf{x}]$) for different training configurations of VCP. \\ *As a reference, we include using the average of the previous $N$ frames as the predicted feature, \ie $\mathbf{\hat{x}}(t) = \sum_{n=1}^{N} \mathbf{x}(t-n)/N$.}
\label{tab:mse}
\end{table}
\fi

\paragraph{Pruning of the candidate boundaries using supervised learning}
For high recall results, false candidate boundaries can be discarded using cluster analysis between the frames that the candidate separates. Having annotated data to train a pruning model can improve the performance of the segmentation algorithm in terms of precision, having minimal impact on the recall. We tested this hypothesis training an SVM model to detect false positives. As can be observed in Table~\ref{tab:results:detailed}, such model improves the average precision of CES by $15\%$ (absolute gain of $10\%$), while recall only decreases by $7.8\%$ (absolute loss of $6\%$). The benefit of using a supervised SVM pruning is much significant for segmentation algorithms of lower precision, such as k-means, even if coming at a higher recall cost.

\paragraph{Performance of CES relative to manual annotations}

Since there is not just one correct way of segmenting video content into events, we have to compare the performance of CES relative to that of the average person. For each lifelog, we average the performance of all available annotations, as evaluated on the selected ground truth. The averaged scores are reported in Table~\ref{tab:results:detailed}.
We observe that subjects are, in f-measure, only $3$ points better than CES. Even though the precision of the manual annotations is very high, the annotators also obtain worse recall than CES. This is due to some of the subjects selecting very general events, \eg wrapping all working afternoon within the same event, disregarding the different meetings. Such annotation criteria yields many false negatives, and therefore drops the recall score. Analogously, in some other cases, subjects selected more details than the ground truth. As a result, their rate of false positives is greater than zero.

CES segments, on average, into more events than the annotators. As a result, it is able to detect $13\%$ more true boundaries than the test subjects, but will also find a relative $70\%$ more incorrect events. Such a large increase is to be expected, as the selected ground truth is very exhaustive, and the annotators rarely identify boundaries not present in the ground truth. Overall, we can conclude that CES is a highly precise event segmentation algorithm. Given our ground truth, CES' f-measure is of $96\%$ relative to the manual performance.

\iftrue
\begin{table}[h]
	\centering
	\setlength\extrarowheight{.2ex}
	\setlength{\tabcolsep}{3pt}
	\begin{tabular}[width=\textwidth]{@{}m{7em}ccc}
		& averaged F1  & averaged Prec. & averaged Rec. \\ \hline\hline
		CES-error & {0.42} & {0.45} & 0.49\\
		CES-mean & 0.52 & 0.56 & 0.56\\
		CES-PCA & 0.66 & 0.67 & 0.69\\
		\textbf{CES (with VCP)} & 0.69 & 0.66 & \textbf{0.77}\\ \hline
		k-means w/ SVM & 0.67 & 0.70 & 0.67\\
		CES  w/ SVM & 0.71 & 0.75 & 0.71\\ \hline
		Manual \mbox{segmentation} & \textbf{0.72} & \textbf{0.80} & 0.68\\
	\end{tabular}
	\caption{Detailed experiments. Comparison of CES with visual context prediction as opposed to using other feature predictions or aggregations; performance of the SVM prunning; and accuracy of the manual annotations against the selected ground truth. (Evaluated on EDUB-Seg20).}
	\label{tab:results:detailed}
\end{table}
\else
\begin{table}[t!]
	\centering
	\setlength\extrarowheight{.2ex}
	\setlength{\tabcolsep}{3pt}
	\begin{tabular}[width=\textwidth]{@{}l|ccc|ccc}
		\multicolumn{1}{@{}c@{}}{}& \multicolumn{3}{@{}c@{}}{EDUB-Seg12}  &  \multicolumn{3}{@{}c@{}}{EDUB-Seg20}\\
		\multicolumn{1}{@{}c@{}|}{method} & F1  & Prec. & Rec. & F1 & Prec. & Rec. \\ \hline
		CES-error & 0.44 & 0.46 & 0.53 & 0.42 & 0.45 & 0.49\\
		CES-mean & 0.51 & 0.53 & 0.58 & 0.52 & 0.56 & 0.56\\
		CES-PCA & 0.66 & 0.67 & 0.69 & 0.66 & 0.67 & 0.69\\
		\textbf{CES (with VCP)} & 0.70 & 0.66 & 0.80 & 0.69 & 0.66 & 0.77\\ 
		\hline
		k-means w/ SVM pruning* & 0.69 & 0.71 & 0.70 & 0.67 & 0.70 & 0.67\\
		CES  w/ SVM pruning* & 0.74 & 0.77 & 0.74 & 0.71 & 0.75 & 0.71\\ 
	\end{tabular}
	\caption{Comparison of CES using other feature predictions and aggregations as opposed to using the visual context prediction.}
	\label{tab:results:detailed}
\end{table}
\fi

\section{Conclusions}
In this paper, we have introduced Contextual Event Segmentation, a novel unsupervised event segmentation method that uses the sequential nature of a photo-stream to infer the presence of event boundaries. At the core of CES is the Visual Context Predictor (VCP), a future sequence generator model that predicts the visual context from a given sequence of frames. The visual context at $t-1$ given the past is compared to that at $t+1$ given the future, to determine whether there is a boundary at frame $t$. 

We have also introduced \emph{R3}, a large scale visual lifelogging dataset depicting a wide variety of events. It is recorded in an unconstrained manner by $57$ independent users, who captured their daily activities morning to evening during over a month. The existence of \emph{R3} has allowed us to train the Visual Context Predictor, which is able to model human activities given sequences of visual features. In a series of experiments, we have proved that the visual context is a strong indicator of event changes. We conjecture that it can also be useful for storytelling and tracking of daily activities.

Leveraging on the visual context of the sequences allows CES to detect boundaries between heterogeneous events and ignore local occlusions and brief diversions. CES improves the performance of the baselines by over $16\%$ in f-measure. The performance of CES is competitive with manual annotations, for which the f-measure is only $3\%$ better than CES'. We propose a fully unsupervised pipeline, which results in greater recall than precision. To improve the precision, supervised pruning can be applied to the final detection step by using cluster consistency analysis. Even though further supervised analysis can be performed to improve that performance, it will always be contingent on the ground truth used, which will be inherently subjective.

\bibliographystyle{ACM-Reference-Format}
\bibliography{biblio}
\end{document}